\pdfoutput=1

\documentclass{article}

\PassOptionsToPackage{numbers}{natbib}
\usepackage[preprint]{neurips_2026}

\usepackage[utf8]{inputenc} %
\usepackage[T1]{fontenc}    %
\usepackage{hyperref}       %
\usepackage{url}            %
\usepackage{booktabs}       %
\usepackage{makecell}       %
\usepackage{wrapfig}        %
\usepackage{amsfonts}       %
\usepackage{nicefrac}       %
\usepackage{microtype}      %
\usepackage[dvipsnames]{xcolor}         %

\newcommand{\numsamples}{237}
\newcommand{\numdynamics}{11}
\newcommand{\name}{VideoFDB\xspace}

\usepackage{amsmath}         %
\usepackage{amssymb}         %
\usepackage{mathtools}       %
\usepackage{bm}              %
\usepackage{graphicx}        %
\usepackage{subcaption}      %
\usepackage{multirow}        %
\usepackage{float}           %
\usepackage{siunitx}         %
\usepackage{xspace}          %
\usepackage[capitalise,noabbrev]{cleveref} %

\usepackage{enumitem}       %
\usepackage{pifont}         %
\usepackage{colortbl}       %
\usepackage{tikz}           %
\usetikzlibrary{arrows.meta,calc,positioning,shapes.multipart} %
\usepackage[breakable,skins,raster]{tcolorbox} %
\usepackage{longtable}      %
\newcommand{\cmark}{\ding{51}}%
\newcommand{\xmark}{\ding{55}}%

\definecolor{insightcolor}{HTML}{BEB085}
\newcounter{insightcounter}

\graphicspath{{fig/}{figures/}{figs/}}

\crefname{section}{Sec.}{Secs.}
\Crefname{section}{Section}{Sections}
\crefname{subsection}{Sec.}{Secs.}
\Crefname{subsection}{Section}{Sections}
\crefname{figure}{Fig.}{Figs.}
\Crefname{figure}{Figure}{Figures}
\crefname{table}{Tab.}{Tabs.}
\Crefname{table}{Table}{Tables}
\crefname{equation}{Eq.}{Eqs.}
\Crefname{equation}{Equation}{Equations}
\crefname{appendix}{App.}{Apps.}
\Crefname{appendix}{Appendix}{Appendices}

\title{\name{}: Evaluating Full-Duplex Vision-Speech Capabilities in Conversational Agents}
\author{
\textbf{Amrita Mazumdar\textsuperscript{\rm 1}}, \
\textbf{Seonwook Park\textsuperscript{\rm 1}}, \
\textbf{Rajarshi Roy\textsuperscript{\rm 1}}, \
\textbf{Nikhil Srihari\textsuperscript{\rm 1}}, \\
\textbf{Shengze Wang\textsuperscript{\rm 1}}, \
\textbf{Yuhao Zhou\textsuperscript{\rm 2}} \
\textbf{Julia Wang\textsuperscript{\rm 2}}, \
\textbf{Koki Nagano\textsuperscript{\rm 1}} \
\textbf{Shalini De Mello\textsuperscript{\rm 1}} \\
\textsuperscript{\rm 1}NVIDIA, \
\textsuperscript{\rm 2}David AI
}
\begin{document}

\maketitle

\begin{abstract}
Natural human conversation is full-duplex and audio-visual: people simultaneously speak and listen while continuously interpreting and producing nonverbal cues, such as nods, smiles, and gestures. 
To support successful human-agent interaction, agents must model full-duplex audiovisual conversation; however, existing full-duplex benchmarks evaluate only speech.
In this work, we present \name{}, the first benchmark to evaluate full-duplex audio-visual-to-audio-visual (AV2AV) conversational agents.
\name{} contributes (i) \numsamples{} dyadic clips spanning \numdynamics{} nonverbal conversational dynamics from real-world video calls, (ii) a taxonomy separating \emph{perception} from \emph{generation} behaviors, and (iii) a rubric-based LM-as-judge evaluation framework with interpretable axes for assessing conversational quality with respect to nonverbal conversational dynamics.
Across open- and closed-source vision-speech agents, we find systematic failure modes: \emph{captioning collapse} and \emph{visual-stream ignorance}, and we show that current systems exploit vision for explicit visual question answering
but not for the streaming joint audiovisual grounding required in natural conversation. 
We further evaluate cascaded speech-to-avatar systems and find that their architecture fundamentally precludes the production of full-duplex nonverbal cues.
As the first benchmark for full-duplex AV2AV interaction, \name{} establishes a foundation for systematic evaluation and, we hope, will accelerate the advancement and development of next-generation multimodal conversational agents.
\end{abstract}

\section{Introduction}
\label{sec:intro}

Advances in multimodal learning and conversational speech agents are transforming how humans interact with AI systems.
As these systems move into embodied settings (e.g., service robots, collaborative agents in telepresence and XR), conversational speech interfaces will proliferate.
Existing speech-to-speech agents can listen and respond to users, but often rely on turn-based interaction~\cite{zhang-etal-2023-speechgpt}.
Natural human conversation, however, is not a sequence of voice turns: it is \emph{full-duplex} and \emph{audio-visual-to-audio-visual} ({AV2AV}), with \textbf{both participants simultaneously speaking, listening, and signaling} through continuous verbal and non-verbal channels~\citep{goodwin1981,devito2019interpersonal}.\footnote{We use the abbreviations AV2AV (audio-visual in/out), AV2A (audio-visual in / audio out — current vision-speech agents), A2A (audio-to-audio), and A2AV (audio in / cascaded avatar out) throughout the paper.}
Such full-duplex, {AV2AV} conversational agents (vision-speech agents) are newly emerging. 
Recent advances, (e.g., Gemini 2.5/3.1 Live~\cite{comanici2025gemini,google_gemini_live_api}, OpenAI Realtime~\cite{openai_realtime_api}, Qwen3-Omni~\cite{xu2025qwen3}, MoshiVis~\cite{royer2025vision}) allow agents to perceive image or low-frame-rate video inputs ({AV2A}) while conversing with users interactively.
Building agents that approach human-like interaction therefore requires (1) systems that simultaneously perceive and generate audio-visual cues continuously, without waiting for explicit turn boundaries, and (2) benchmarks that evaluate their multimodal conversational dynamics, not just the semantic correctness of their responses.

However, no benchmark evaluates the full capabilities of full-duplex vision-speech agents.
Existing full-duplex \emph{speech-to-speech} benchmarks~\citep{peng2025fdbenchfullduplexbenchmarkingpipeline,lin2025fdb_v1,lin2025fdb_v15,lin2026fdb_v2} measure turn-taking, backchanneling, and interruption from audio alone, leaving gesture, gaze, and affect unmeasured.
Conversely, audio-visual VLM benchmarks mostly evaluate \emph{video question answering}~\citep{fu2025video,antol2015vqa,chaubey2026avere,chen2025savvy,qin2025face,xun2025rtv,tang2025mmperspective}, rewarding semantic correctness over conversational dynamics, while audio-driven avatar benchmarks~\citep{peng2025dualtalk,liu2024emage,liu2022beat} evaluate speaker and listener behaviors separately rather than via continuous joint interaction. Across these domains, no work evaluates the agent as an active interlocutor (i.e., conversational participant) that must continuously engage with human nonverbal dynamics.

Evaluation is intrinsically difficult because conversation is non-deterministic: the same user signal (e.g., laughter) can merit multiple valid responses depending on  context and interaction style~\cite{Watzlawick1967Pragmatics}.
Correct assessment must therefore allow a distribution of valid behaviors, rather than a single correct response.
To our knowledge, no publicly available full-duplex AV2AV systems exist, underscoring the need for further research.

\begin{figure}[t]
  \centering
  \includegraphics[width=\textwidth]{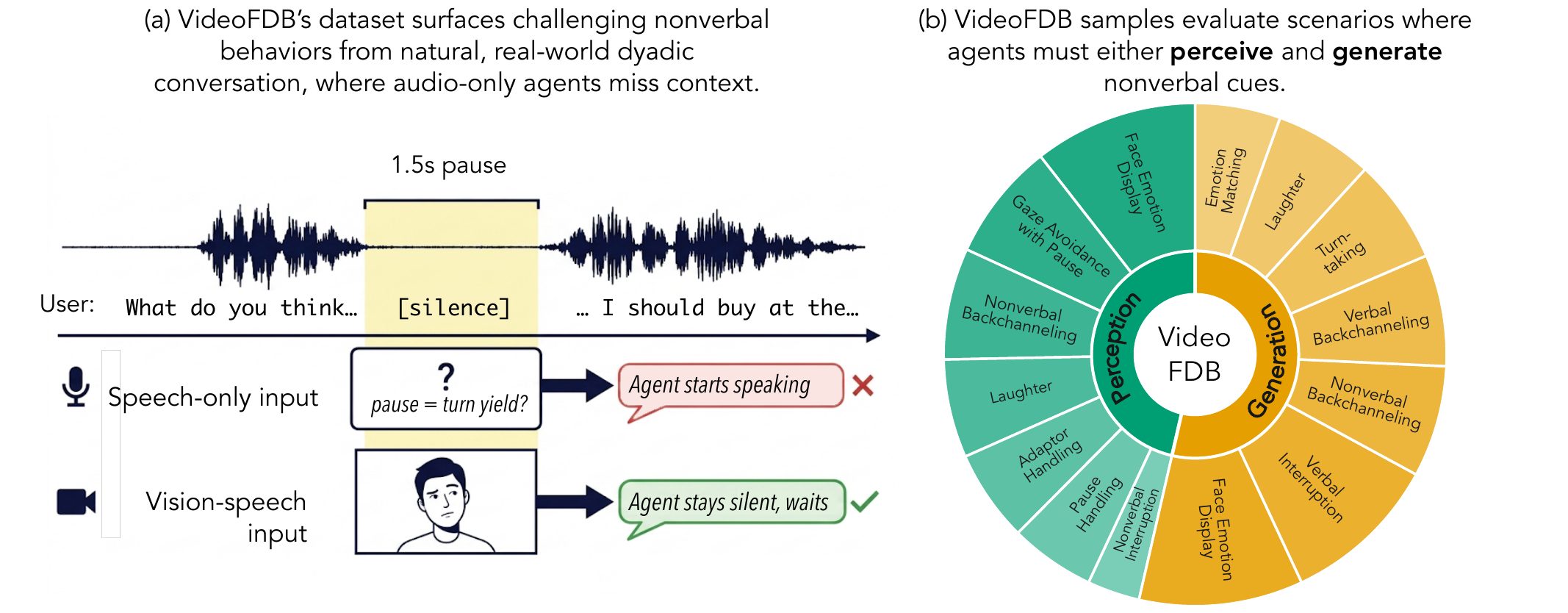}
  \caption{\name{} (a) curates evaluation samples from natural conversations, where visual cues carry key context, and (b) evaluates perception and generation capabilities of full-duplex audio-visual conversational agents across \numdynamics~categories, including dynamics that appear in both tracks.}
  \label{fig:overview}
  \vspace{-1em}
\end{figure}

In this paper, we present \name{}, the first benchmark evaluating the conversational dynamics of full-duplex audio-visual conversational agents, grounded in a taxonomy of nonverbal dynamics~\citep{goodwin1981,devito2019interpersonal}.
\name{} treats the agent as an active conversational participant and evaluates responses along axes in two categories:
\begin{itemize}[leftmargin=*]
\setlength\parskip{0em}
\item \textbf{Perception}: (1) \textit{Fluency}: whether the interaction remains coherent and conversationally natural; (2) \textit{Conversational Flow}: whether the agent can successfully perceive nonverbal cues when managing turn timing and floor transitions appropriately (e.g., yielding, holding, interruption, and backchannel timing); and (3) \textit{Semantic Grounding}: whether the agent's response behavior is semantically aligned with perceived nonverbal cues.
\item \textbf{Generation} (if agent video is produced): (1) \textit{Fluency}, as in the Perception category; (2) \textit{Dyadic Affect Match}: whether the agent's combined AV response affectively corresponds with the user's affective state; and (3) \textit{Nonverbal Cue Appropriateness}: whether produced nonverbal behaviors are category-appropriate and produced at appropriate moments in the conversation.
\end{itemize}
\name{} includes \numdynamics~conversational dynamics with \numsamples~human-annotated audio-visual dyadic samples (\cref{fig:overview}), with evaluation protocols for both categories. We systematically evaluate leading open-source and proprietary full-duplex vision-speech agents along with cascaded speech-to-avatar systems, finding that current systems miss nonverbal cues, fall back to captioning-style replies, and degrade speech quality as user video sampling increases. Further, we observe that AV2A vs.\ A2A comparisons indicate visual input is used for captioning but ignored when visuals are not verbally referenced. Finally, we find cascaded speech-to-avatar pipelines preserve turn-yielding discipline but cannot insert cues during the user's turn, with latencies $2.8$--$3.5$\,s behind human ground truth.

To summarize, our main contributions are:
\begin{itemize}[leftmargin=*]
 \setlength\parskip{0em}
\item We introduce \name{}, the first full-duplex vision-speech benchmark for jointly evaluating verbal and nonverbal conversational dynamics in natural dyadic conversation.
\item We benchmark both open-source and proprietary full-duplex conversational speech agents and show that even state-of-the-art systems struggle to perceive and respond to natural nonverbal dynamics.
\item We provide a failure-mode analysis that identifies \emph{captioning collapse}, \emph{visual-stream ignorance}, and \emph{structural challenges in cascaded speech-avatar systems}, clarifying key directions towards advancing end-to-end audio-visual conversational agents.

\end{itemize} %
\section{Related Work}
\label{sec:related-work}

\paragraph{Full-duplex speech agents and omni-models.}
Full-duplex speech-to-speech methods natively model both sides of a dyadic exchange in a single network: Moshi~\cite{defossez2024moshi} and dGSLM~\cite{nguyen2023generative} jointly generate user and agent streams to support overlapping speech, with OmniFlatten~\cite{zhang-etal-2025-omniflatten}, SyncLLM~\cite{veluri-etal-2024-beyond}, SALM~\cite{hu2025salm}, and PersonaPlex~\cite{roy2026personaplexvoicerolecontrol} extending this template to richer turn-taking and role control.
Recent work extends full-duplex speech-to-speech models: Gemini 2.5 and 3.1 Flash Live~\cite{google_gemini_live_api, comanici2025gemini}, GPT Realtime~\citep{openai_realtime_api, hurst2024gpt}, Qwen2.5- and Qwen3-Omni~\citep{qwen2025omni, xu2025qwen3}, and MoshiVis~\cite{royer2025vision} all accept audio and video inputs jointly while preserving streaming speech output, and Video-SALMONN~\cite{sun2024videosalmonn} adds video to streaming dialogue understanding.
OmniResponse~\cite{luo2025omniresponse} extends multimodal language models (MLLM) to model dyadic conversation, including verbal and nonverbal cues; their method generates full-duplex audio-visual conversational behavior, but is only evaluated on uni-modal speech, visual, or text quality.
None of these systems, to our knowledge, has been evaluated on its ability to interact with the nonverbal component of a full-duplex exchange, such as gesture, gaze, or facial signal.

\begin{wraptable}[10]{r}{0.55\textwidth}
    \centering
    \vskip -3mm
    \caption{Prior benchmarks isolate speech-to-speech interaction or turn-based/split-role vision-speech tasks.}
    \vskip -1mm
    \resizebox{\linewidth}{!}{%
    \begin{tabular}{l c c c c c c}
        \toprule
        \textbf{\shortstack{\\Benchmark}} & \textbf{\shortstack{Full-\\duplex?}} & \textbf{\shortstack{Speech\\In?}} & \textbf{\shortstack{Speech\\Out?}} & \textbf{\shortstack{Video\\In?}} & \textbf{\shortstack{Video\\Out?}} & \textbf{\shortstack{Real-\\world?}} \\
        \midrule
        \rowcolor{gray!10}
        Full-Duplex Bench~\cite{lin2025fdb_v1}
            & \textcolor{green!60!black}{\cmark}
            & \textcolor{green!60!black}{\cmark}
            & \textcolor{green!60!black}{\cmark}
            & \textcolor{red!70!black}{\xmark}
            & \textcolor{red!70!black}{\xmark}
            & \textcolor{red!70!black}{\xmark} \\
        EMO-Reasoning~\citep{liu2025emoreasoning}
            & \textcolor{red!70!black}{\xmark}
            & \textcolor{green!60!black}{\cmark}
            & \textcolor{green!60!black}{\cmark}
            & \textcolor{red!70!black}{\xmark}
            & \textcolor{red!70!black}{\xmark}
            & \textcolor{red!70!black}{\xmark} \\
        \rowcolor{gray!10}
        EmoRealm~\citep{chaubey2026avere}
            & \textcolor{red!70!black}{\xmark}
            & \textcolor{red!70!black}{\xmark}
            & \textcolor{red!70!black}{\xmark}
            & \textcolor{green!60!black}{\cmark}
            & \textcolor{red!70!black}{\xmark}
            & \textcolor{green!60!black}{\cmark} \\
        OmniMMI~\citep{omnimmi}
            & \textcolor{red!70!black}{\xmark}
            & \textcolor{red!70!black}{\xmark}
            & \textcolor{red!70!black}{\xmark}
            & \textcolor{green!60!black}{\cmark}
            & \textcolor{red!70!black}{\xmark}
            & \textcolor{green!60!black}{\cmark} \\
        \rowcolor{gray!10}
        ResponseNet~\citep{luo2025omniresponse}
            & \textcolor{red!70!black}{\xmark}
            & \textcolor{green!60!black}{\cmark}
            & \textcolor{green!60!black}{\cmark}
            & \textcolor{green!60!black}{\cmark}
            & \textcolor{green!60!black}{\cmark}
            & \textcolor{green!60!black}{\cmark} \\
        \midrule
        \textbf{\name~ (Ours)}
            & \textcolor{green!60!black}{\cmark}
            & \textcolor{green!60!black}{\cmark}
            & \textcolor{green!60!black}{\cmark}
            & \textcolor{green!60!black}{\cmark}
            & \textcolor{green!60!black}{\cmark}
            & \textcolor{green!60!black}{\cmark} \\
        \bottomrule
    \end{tabular}}
    \label{tab:problem-setting-comparison}
\end{wraptable}

\paragraph{Multimodal conversational benchmarks.}
Existing multimodal evaluations probe audio-visual understanding through paired prompts and text responses rather than continuous interaction.
Captioning~\cite{fu2025video} and visual question answering (VQA)~\cite{antol2015vqa} methods directly evaluate visual understanding; AVERE~\cite{chaubey2026avere}, SAVVY~\cite{chen2025savvy}, and MMPerspective~\cite{tang2025mmperspective} extend these evaluations to richer audio-visual reasoning, and recent methods target face-centric\cite{qin2025face}, real-time~\cite{xun2025rtv}, and situated~\cite{pourreza2026can} VQA, where the model answers questions about the visible person or scene rather than participating as that person's conversational partner.

A complementary line~\cite{chen2025savvy, radevski2025dave} specifically curates VQA pairs that probe cross-modal integration, but their evaluations are also turn-based.

Multimodal evaluation methods that focus on dyadic interaction (e.g., speaker or listener generation) commonly use \emph{split-role} protocols that bifurcate the exchange into discrete speaker and listener roles: OmniMMI~\cite{omnimmi} measures proactive turn-taking against scripted prompts,~\cite{nguyen2025see} evaluates turn-based grounded visual reference resolution, and ResponseNet~\cite{luo2025omniresponse} scores agent listener behavior in dyadic conversation by separating speaker and listener turns rather than evaluating cross-role overlap.
In these split-role settings, models are evaluated as either \emph{speaker} or \emph{listener} at a time, not as interlocutors continuously managing both speaker and listener turns in overlapping conversation.
Across multimodal evaluation methods, success metrics score per-turn response quality, and, optionally, latency. No method evaluates continuous conversation or naturalness under full-duplex audio-visual settings.

\paragraph{Full-duplex speech evaluation.}
Full-duplex speech benchmarks such as Full-Duplex-Bench~\cite{peng2025fdbenchfullduplexbenchmarkingpipeline, lin2025fdb_v1, lin2025fdb_v15, lin2026fdb_v2} measure dynamics like turn-taking, interruption handling, and backchannel timing.
EMO-Reasoning~\cite{liu2025emoreasoning} evaluates emotion reasoning in speech without modeling full-duplex dynamics.
\paragraph{Avatar animation evaluation.} 
Cascaded audio-driven avatar methods, where generated speech drives portrait animation in real time, are typically evaluated on visual realism and naturalness~\cite{chen2020talking,ding2025kling,xing2023codetalker}.
Benchmarks evaluating audio-driven gesture generation~\cite{peng2025dualtalk,liu2024emage,liu2022beat} also focus on motion realism and gesture fidelity in isolation of conversational semantics.

\name is, to our knowledge, the first benchmark to evaluate interaction-level AV2AV dynamics---gesture, gaze, facial signal, and dialogue cues---in genuinely full-duplex conversation, by: (1) evaluating overlapping dyadic exchange with role switching; (2) covering a broad range of vision-speech dynamics; and (3) scoring interactional appropriateness (perception and generation) rather than only semantic correctness or uni-modal generation quality (\Cref{tab:problem-setting-comparison}).

\section{\name~ Benchmark Dataset}
\label{sec:benchmark}

\name~ evaluates the visual perception and visual generation capabilities of a conversational speech agent in a natural, full-duplex conversation with a benchmark dataset of conversational dynamics.
\name{}'s dataset includes dynamics identified in human communication research as necessary for fluent dyadic interaction~\cite{devito2019interpersonal}, including dialogue-management cues (gaze, intonation, gesture completion), backchannels, interruption signals, affect displays, and body movements.
This grounding ensures that \name{} measures competence on dynamics human interlocutors \emph{actually rely on}, rather than arbitrary visual phenomena or direct visual question-answering.

\begin{table*}[t]
  \centering
  \small
  \renewcommand{\arraystretch}{1.2}
  \caption{\name{} conversational dynamics overview.}
  \vskip -1mm
  \resizebox{\textwidth}{!}{
    \begin{tabular}{ccccp{0.25\textwidth}p{0.50\textwidth}cc}
    \toprule
    \multicolumn{4}{c}{\textbf{Nonverbal Communication Channels}} &
    \multirow{2}{*}{\textbf{\shortstack{Conversational\\Dynamic}}} &
    \multirow{2}{*}{\textbf{Dynamic Description}} &
    \multicolumn{2}{c}{\textbf{Evaluation Category}} \\
    \cmidrule(lr){1-4} \cmidrule(lr){7-8}
    \textbf{Dialogue} & \textbf{Eye Gaze} & \textbf{Face} & \textbf{Body} &  &  & \textbf{Perception} & \textbf{Generation} \\
    \midrule
    \multicolumn{8}{l}{\scriptsize\textit{Perception-only dynamics}} \\ \midrule
    \textcolor{green!60!black}{\cmark} & \textcolor{green!60!black}{\cmark} &  &  & Gaze Avoidance with Pause & Gaze aversion paired with a pause often indicating thinking or processing. & \textcolor{green!60!black}{\cmark} &  \\
    &  &  & \textcolor{green!60!black}{\cmark} & Adaptor Handling & Self-directed or reflexive action, such as coughing, yawning, scratching, or adjusting hair. & \textcolor{green!60!black}{\cmark} &  \\
    \textcolor{green!60!black}{\cmark} & \textcolor{green!60!black}{\cmark} &  &  & Pause Handling & Brief pause within a speaker's turn, typically for thinking or a small action. & \textcolor{green!60!black}{\cmark} &  \\
    \textcolor{green!60!black}{\cmark} &  &  & \textcolor{green!60!black}{\cmark} & Nonverbal Interruption & Interruption through gesture, facial expression, or other nonverbal behavior, optionally paired with speech. & \textcolor{green!60!black}{\cmark} &  \\
    \addlinespace
    \midrule
    \multicolumn{8}{l}{\scriptsize\textit{Shared dynamics (Perception + Generation)}} \\ \midrule
    & \textcolor{green!60!black}{\cmark} & \textcolor{green!60!black}{\cmark} &  & Face Emotion Display & Visible facial emotional expression during the interaction. & \textcolor{green!60!black}{\cmark} & \textcolor{green!60!black}{\cmark} \\
    \textcolor{green!60!black}{\cmark} &  & \textcolor{green!60!black}{\cmark} &  & Nonverbal Backchanneling & Listener feedback delivered through facial expression, sometimes paired with speech. & \textcolor{green!60!black}{\cmark} & \textcolor{green!60!black}{\cmark} \\
    \textcolor{green!60!black}{\cmark} &  & \textcolor{green!60!black}{\cmark} &  & Laughter & Laughter during the interaction. & \textcolor{green!60!black}{\cmark} & \textcolor{green!60!black}{\cmark} \\
    \addlinespace
    \midrule
    \multicolumn{8}{l}{\scriptsize\textit{Generation-only dynamics}} \\ \midrule
    \textcolor{green!60!black}{\cmark} &  &  &  & Verbal Interruption & Spoken interruption while another participant is still talking. &  & \textcolor{green!60!black}{\cmark} \\
    \textcolor{green!60!black}{\cmark} &  & \textcolor{green!60!black}{\cmark}   &  & Verbal Backchanneling & Short listener feedback delivered through speech, sometimes paired with nonverbal behavior. &  & \textcolor{green!60!black}{\cmark} \\
    \textcolor{green!60!black}{\cmark} &  &  &  & Turn-taking & Exchange of speaking roles between participants. &  & \textcolor{green!60!black}{\cmark} \\
    &  & \textcolor{green!60!black}{\cmark} &  & Emotion Matching & Mirroring of the speaker's emotional expression by the listener. &  & \textcolor{green!60!black}{\cmark} \\
    \bottomrule
  \end{tabular}
  }
  \label{tab:dynamics-overview}
\end{table*}

\Cref{sec:benchmark-dynamics} presents an overview of the conversational dynamics included in our dataset and what they evaluate, and \Cref{sec:benchmark-stats} summarizes dataset statistics and the capture/annotation pipeline.

\subsection{\name~Conversational Dynamics Taxonomy}
\label{sec:benchmark-dynamics}

We organize our benchmark around a taxonomy of nonverbal channels that human communication science identifies as central to how interlocutors co-construct meaning~\citep{goodwin1981,devito2019interpersonal,Watzlawick1967Pragmatics}:
\begin{itemize}[leftmargin=*]
  \item \textbf{Dialogue}: While dialogue tends to refer to verbal dialogue cues, conversations also include paralinguistic or nonverbal dialogue cues: turn-taking depends on timing and backchannel behavior; longer gaps often carry pragmatic meaning; and acknowledgment tokens (e.g., \emph{mm-hm}, \emph{yeah}) mark attentive listening without claiming the floor~\citep{levinson2015timing,meyer2020timing,Yngve1970OnGA}
  \item \textbf{Eye Gaze}: Beyond lexical content, gaze behavior contributes independent conversational information. Gaze direction and aversion regulate turn transitions and indicate social intent (e.g., glancing away may signal turn-yielding or a mid-thought pause)~\citep{devito2019interpersonal,goodwin1981, Watzlawick1967Pragmatics}
  \item \textbf{Face} and \textbf{Body}: Beyond spoken words, facial and bodily behavior carry additional conversational signals. Co-speech gestures (movements produced synchronously with speech), adaptors (self-directed actions like fidgeting), and facial affect displays provide complementary semantic and interpersonal cues that shape how utterances are interpreted~\citep{morett2024examining,alibali2001effects,devito2019interpersonal}
\end{itemize}

We operationalize this communication framework by defining, for each nonverbal channel, conversational dynamics that are observable, annotatable, and behaviorally relevant in dyadic interaction.
Specifically, we select dynamics that directly govern conversational floor management and turn signaling in natural exchange.
\Cref{tab:dynamics-overview} describes the dynamics, how they map to nonverbal communication channels and indicates whether each dynamic is used in the \textbf{Perception} and/or \textbf{Generation} evaluations.
Across the taxonomy, the selected dynamics cover floor management, listener feedback, social-affective signaling, and conversational body movement.

\subsection{Dataset Statistics, Capture, and Annotation}
\label{sec:benchmark-stats}

\begin{figure}[t]
  \centering
  \vskip -2mm
  \includegraphics[width=0.98\textwidth]{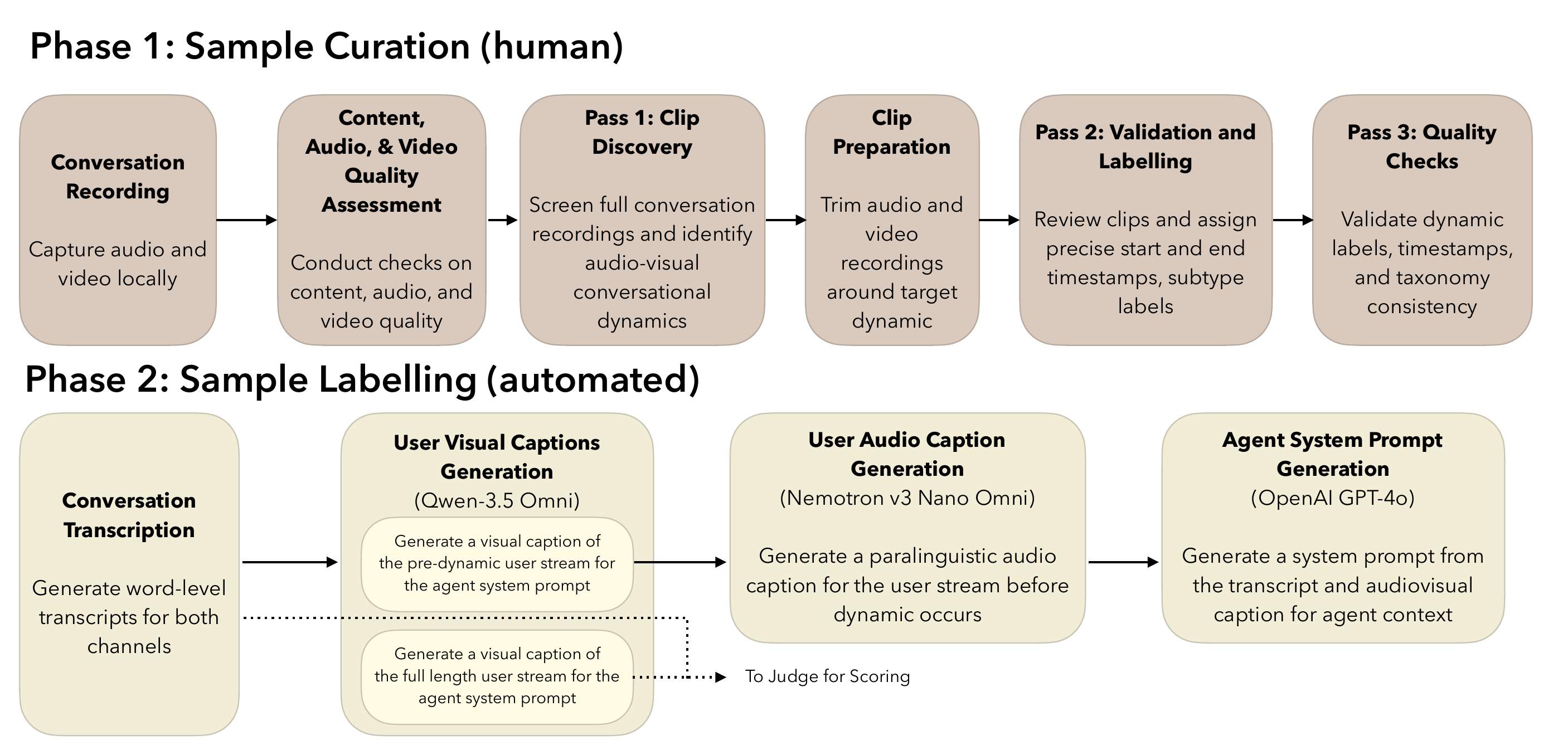}
  \vskip -1mm
  \caption{Annotation workflow. Human annotators select and curate clips over multiple passes of quality reviews. Automated captioning generates contextual prompts for agent inference and \name{} evaluation.}
  \vskip -3mm
  \label{fig:annotation-pipeline}
\end{figure}

\name~ contributes \numsamples~clips from natural two-person video calls, spanning \numdynamics~dynamic types.
Each sample contains diarized, speaker-separated audio/video streams for both participants and is centered on one annotated \textbf{conversational dynamic} with event window $[t_\text{start}, t_\text{end}]$, while preserving salient surrounding context (typically 1--3 lead-in turns and up to one follow-up turn).
To support evaluation, recordings are filtered and manually annotated into a standardized label set.
\Cref{fig:annotation-pipeline} illustrates the end-to-end annotation process. 
Appendix~\ref{sec:appendix-dataset-collection} provides additional dataset details.

\paragraph{Collection and annotation.}
Our dataset is mined from a held-out corpus of two-person video calls, in which participants converse naturally over a videoconferencing platform, with each speaker's stream recorded locally to mitigate network-latency artifacts.
Source recordings are then screened with media-quality and human preprocessing checks.
Final clips are curated with a three-pass human annotation pipeline (candidate discovery, timestamp/type validation with sufficient pre-event context, and final quality review). Appendix~\ref{sec:appendix-dataset-collection} provides full data curation details. 

After selection and annotation, each clip is supplemented with LM-generated labels: a system prompt for agent bootstrapping, a user-stream AV caption for judge context, and an event-window dynamic label. \Cref{subsec:system-prompt-preparation} describes the system prompt and AV caption pipelines. 
\section{Evaluation \& Experimental Setup}
\label{sec:eval-protocol}

Natural conversation is fundamentally non-deterministic: for most nonverbal events, there is no single ``correct'' response string. 
\name~ thus uses rubric-based LM-as-judge~\cite{zheng2023judging} scoring instead of exact-match targets, which allows us to rate multiple semantically valid responses as acceptable and score response quality on interpretable dimensions. 
In this section, we describe \name{}'s evaluation protocol and scoring metrics.

\subsection{Agent Perception \& Generation Evaluations}
We group conversational dynamics into two categories: \emph{perception} dynamics test whether agents correctly interpret user-produced nonverbal behavior, while \emph{generation} dynamics test whether agents produce appropriate nonverbal responses (only applicable to full-duplex agents that emit continuous visual output).
We evaluate visual perception against three interpretable rubrics: \textbf{Fluency} grades overall interaction quality (talk-over, monologue, nonsensical responses); \textbf{Conversational Flow} grades timing of agent responses around a nonverbal cue; \textbf{Semantic Grounding} grades response content against visual-affective event semantics. The latter two apply only to relevant dynamics: Pause Handling rewards \emph{not} taking the floor while Backchanneling rewards \emph{continuing} (both under Flow); Nonverbal Interruption is scored along both axes (yield timing under Flow, post-yield behavior under Grounding).

For agents generating visual output, we evaluate: \textbf{Fluency} (as above); \textbf{Dyadic Affect Match}, whether the agent's combined audio+visual response affectively corresponds with the user's affective state; and \textbf{Nonverbal Cue Appropriateness}, whether produced nonverbal behaviors are category-appropriate and well-timed. \Cref{tab:dynamic-axis-map} gives the full dynamic-by-axis breakdown.

\begin{figure}[t]
  \centering
  \vskip -2mm
  \includegraphics[width=0.98\textwidth]{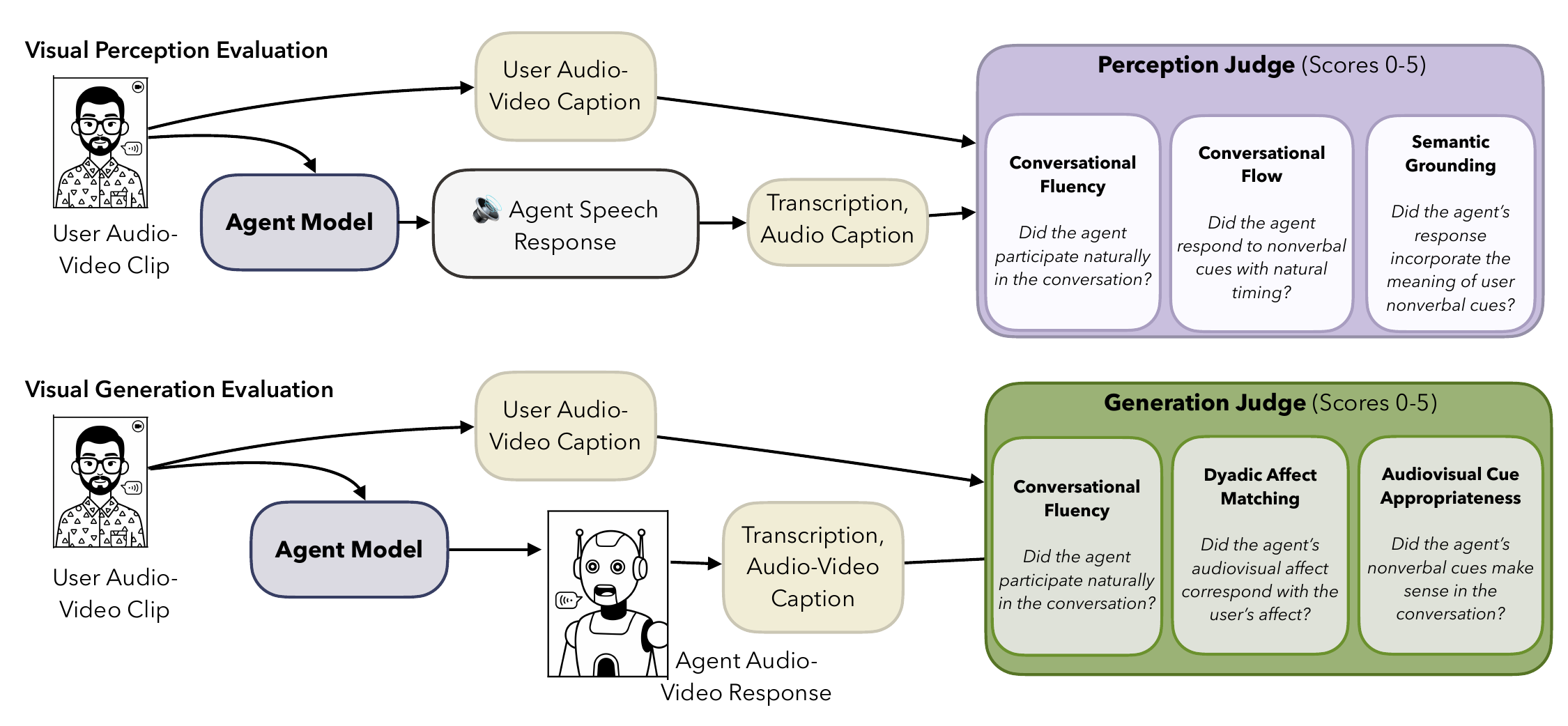}
  \vskip -2mm
  \caption{Evaluation flow. We feed agent-generated audio and/or video with annotated captions
           into judge prompts, returning category-specific scores.}
  \vskip -4mm
  \label{fig:evaluation-flow}
\end{figure}

\subsection{Evaluation Scoring \& Metrics}
We now describe how we evaluate agent responses to our benchmark samples using \name{}'s rubrics and metrics. 
The end-to-end evaluation protocol is summarized in \Cref{fig:evaluation-flow}.
For each clip, we supply a fixed system prompt and then stream the user audio/video channel to the agent in real time, and record the agent output stream.
The recorded agent stream is transcribed with word-level timestamps, then merged with clip metadata (e.g., \texttt{dynamic\_type}, \texttt{dynamic\_start\_s}/\texttt{dynamic\_end\_s}, AV caption) to construct the judge payload.
The judge then scores each clip on the configured rubric axes and computes the deterministic timing metrics as follows.

\paragraph{Scoring ``Appropriate'' Conversational Behavior} 
We score appropriateness with pre-specified, dynamic-specific rubric anchors.
For each dynamic, the rubric defines the expected interaction policy around the annotated event window (e.g., \textit{stay silent}, \textit{continue speaking}, \textit{yield}, \textit{affectively align}) and maps clip evidence (timestamps, transcripts, and AV caption) to a shared 0--5 scale.
Appendix~\ref{sec:judge-prompts} details full prompts and rubrics.

\paragraph{Timing Metrics.}\label{par:timing-metrics}
In addition to the rubric axes above that capture timing quality qualitatively, we also score timing behavior directly, leveraging our dynamic window annotations and the generated agent response stream's timestamps. 
To report unified timing scores across nonverbal cues with different expected turn-taking outcomes, we introduce \textbf{Takeover-Rate Alignment} (TOR-Alignment). 
TOR-Alignment extends the speech-only TOR formulation of \citet{lin2025fdb_v1} to multimodal AV dynamics by unifying heterogeneous timing expectations under one metric; higher TOR-Alignment indicates a larger fraction of samples satisfying expected timing behavior.
We map timing-relevant dynamics to one of five timing classes, defined by the expected agent behavior when the dynamic appears: \textsc{stay-silent}, \textsc{continue-speaking}, \textsc{yield-required}, \textsc{smooth-handoff}, and \textsc{backchannel-produced}. 
Appendix~\ref{sec:appendix-timing-metrics} provides a full formalization.

\textbf{Latency.} We also measure the latency of the agent's response to the timing class, which distinguishes \emph{stay-silent} from \emph{continue-speaking}: both can involve short silences, but the expected behavior differs by conversational role (listener vs.\ current speaker).

\subsection{Additional Annotations}
\paragraph{System prompt preparation for agent evaluation.}
\label{subsec:system-prompt-preparation}
When a user event occurs amid an ongoing agent turn (e.g., nonverbal interruption), we condition the agent with a short system prompt summarizing pre-event context and explicitly triggering speech (``Start speaking now to begin/continue the conversation''). Appendix~\ref{sec:judge-prompts-system-prompt} provides additional details.

\paragraph{Judge prompt construction \& validation}
To provide agent responses to our judge for scoring, we construct a structured judge payload (event metadata, role-labeled transcripts, and AV captions), using Qwen3.5-397B~\cite{qwen35} for visual captions and Nemotron v3 Nano Omni 30B~\cite{nvidia2026nemotron3nanoomni} for audio captions; Appendix~\ref{sec:judge-prompts} provides full details.
Our rubrics show stable cross-judge consistency: across three judge models, pairwise agreement is $77$--$89\%$ within $1$ point on the $0$--$5$ scale (Appendix~\ref{sec:appendix-judge-agreement}).
Following recent full-duplex speech evaluations~\cite{lin2025fdb_v1, roy2026personaplexvoicerolecontrol}, we employ \texttt{gpt-4o} for scoring.

\section{Experiments and Results}
\label{sec:results}

We evaluate \name~ on a diverse set of full-duplex models, spanning speech-vision, audio-only, and speech-to-avatar systems.

\noindent \textbf{Closed- and open-source vision-speech models (AV2A).} Our closed-source speech-vision baselines are Gemini~2.5 and 3.1~Flash Native \cite{comanici2025gemini} and OpenAI Realtime and Realtime mini~\cite{openai_realtime_api}. Our open-source speech-vision baselines are MiniCPM-o~4.5 \cite{minicpmo2025}, MiniOmni2 \cite{xie2024miniomni2}, and VITA-1.5 \cite{fu2026vita}. MiniOmni2 only accepts one frame per user speech segment; for all other models, we stream user video at 1 FPS. We stream audio input at 16-24 kHz, depending on model requirements.

\noindent \textbf{Closed- and open-source conversational speech models (A2A).} To isolate the contribution of visual cues, we re-run all vision-speech agents in audio-only mode, comparing paired A2A and AV2A runs on the same clips.

\noindent \textbf{Full-duplex speech models with cascaded avatar generation.} Because no publicly available full-duplex AV2AV models exist, we evaluate a cascaded setup where a full-duplex conversational speech model (AV2A) runs in LiveKit Cloud~\cite{livekit_cloud} and its speech drives an audio-driven avatar (e.g., Anam~\cite{anam2026}, Keyframe~\cite{keyframe2026}). We use a single full-duplex speech backbone (Gemini 2.5 Flash Native) for both avatars to isolate avatar effects from speech-backbone variation.

\name{} tests whether these systems respond to full-duplex user input with temporally aligned, affect-consistent, and category-appropriate visible conversational behavior (\Cref{tab:leaderboard-perception,tab:leaderboard-generation}).

\begin{table*}[t]
  \centering
  \caption{Performance breakdown across Perception rubrics for full-duplex speech-vision models. Timing reports TOR-Alignment percentage above and median latency below. Best and second-best non-human entries are marked in bold and underline, respectively.}
  \label{tab:leaderboard-perception}
  \small
  \renewcommand{\arraystretch}{1.2}
  \resizebox{.7\textwidth}{!}{
\begin{tabular}{l | c c c >{\columncolor{gray!15}}c | c}
    \toprule
  \textbf{Model} &
    \textbf{Fluency $\uparrow$} & \textbf{Conv.~Flow $\uparrow$} & \textbf{Vis.~Ground.\ $\uparrow$} & \textbf{Overall $\uparrow$} & \textbf{Timing $\uparrow$} \\
   \midrule
 Human reference & 4.16 {\scriptsize (3.88--4.42)} & 4.20 {\scriptsize (3.89--4.50)} & 4.24 {\scriptsize (3.95--4.47)} & 4.20 & \makecell{90\% \\ {\scriptsize 1400\,ms}} \\
   \midrule
  \multicolumn{6}{l}{\textit{Closed-source full-duplex speech-vision models}} \\
   \midrule
 Gemini 2.5 Flash Native & 3.33 {\scriptsize (2.91--3.75)} & 2.81 {\scriptsize (2.20--3.43)} & 3.37 {\scriptsize (2.97--3.78)} & 3.17 & \makecell{72\% \\ {\scriptsize 3160\,ms}} \\
 Gemini 3.1 Flash Live & 3.15 {\scriptsize (2.74--3.55)} & 2.20 {\scriptsize (1.62--2.79)} & 3.16 {\scriptsize (2.52--3.74)} & 2.84 & \makecell{66\% \\ {\scriptsize 1720\,ms}} \\
 OpenAI \texttt{gpt-realtime-mini} & 2.91 {\scriptsize (2.48--3.34)} & 2.37 {\scriptsize (1.76--3.02)} & 2.90 {\scriptsize (2.42--3.36)} & 2.73 & \makecell{66\% \\ {\scriptsize 5320\,ms}} \\
 OpenAI \texttt{gpt-realtime} & 2.72 {\scriptsize (2.27--3.16)} & 2.50 {\scriptsize (1.91--3.09)} & 3.02 {\scriptsize (2.53--3.49)} & 2.75 & \makecell{72\% \\ {\scriptsize 5400\,ms}} \\
   \midrule
  \multicolumn{6}{l}{\textit{Open-source full-duplex speech-vision models}} \\
   \midrule
MiniCPM-o 4.5~\cite{minicpmo2025} & 3.03 {\scriptsize (2.64--3.39)} & \underline{3.54} {\scriptsize (3.02--4.04)} & \textbf{3.63} {\scriptsize (3.25--4.00)} & \underline{3.40} & \makecell{\textbf{73\%} \\ {\scriptsize \textbf{720\,ms}}} \\
 MiniOmni2~\cite{xie2024miniomni2} & 0.65 {\scriptsize (0.42--0.90)} & 1.37 {\scriptsize (0.87--1.91)} & 1.54 {\scriptsize (1.08--2.03)} & 1.19 & \makecell{64\% \\ {\scriptsize 3080\,ms}} \\
 VITA-1.5~\cite{fu2026vita} & 1.19 {\scriptsize (0.94--1.45)} & 1.57 {\scriptsize (1.02--2.11)} & 2.53 {\scriptsize (2.10--2.98)} & 1.76 & \makecell{58\% \\ {\scriptsize 400\,ms}} \\
  \midrule
  \multicolumn{6}{l}{\textit{Audio-only full-duplex speech-vision models}} \\
   \midrule
Gemini 2.5 Flash Native & 3.35 {\scriptsize (2.95--3.78)} & 2.98 {\scriptsize (2.41--3.61)} & 3.17 {\scriptsize (2.69--3.63)} & 3.17 & \makecell{\underline{73\%} \\ {\scriptsize \underline{2760\,ms}}} \\
Gemini 3.1 Flash Live & \underline{3.40} {\scriptsize (3.01--3.77)} & 2.64 {\scriptsize (2.07--3.21)} & 3.03 {\scriptsize (2.32--3.71)} & 3.03 & \makecell{69\% \\ {\scriptsize 1240\,ms}} \\
 OpenAI \texttt{gpt-realtime-mini} & 3.05 {\scriptsize (2.62--3.47)} & 2.48 {\scriptsize (1.91--3.07)} & 3.12 {\scriptsize (2.71--3.54)} & 2.88 & \makecell{69\% \\ {\scriptsize 5000\,ms}} \\
OpenAI \texttt{gpt-realtime} & 2.93 {\scriptsize (2.50--3.39)} & 2.37 {\scriptsize (1.78--2.96)} & \underline{3.59} {\scriptsize (3.17--4.00)} & 2.97 & \makecell{67\% \\ {\scriptsize 4440\,ms}} \\
MiniCPM-o 4.5~\cite{minicpmo2025} & \textbf{3.45} {\scriptsize (3.08--3.79)} & \textbf{3.76} {\scriptsize (3.31--4.19)} & 3.10 {\scriptsize (2.59--3.59)} & \textbf{3.44} & \makecell{72\% \\ {\scriptsize 920\,ms}} \\
 MiniOmni2~\cite{xie2024miniomni2} & 1.48 {\scriptsize (1.13--1.84)} & 1.70 {\scriptsize (1.13--2.28)} & 2.15 {\scriptsize (1.69--2.63)} & 1.72 & \makecell{69\% \\ {\scriptsize 2760\,ms}} \\
 VITA-1.5~\cite{fu2026vita} & 1.62 {\scriptsize (1.33--1.90)} & 1.37 {\scriptsize (0.87--1.89)} & 3.02 {\scriptsize (2.61--3.42)} & 2.00 & \makecell{61\% \\ {\scriptsize 800\,ms}} \\

   \bottomrule
 \end{tabular}
 }
\end{table*}

\begin{table*}[t]
  \centering
  \caption{Performance breakdown across Generation rubrics for full-duplex speech-vision models with cascaded avatars. Timing reports TOR-Alignment percentage above and median latency below. Best scores in bold.}
  \label{tab:leaderboard-generation}
  \small
  \renewcommand{\arraystretch}{1.2}
  \resizebox{.7\textwidth}{!}{
    \begin{tabular}{l |
      >{\centering\arraybackslash}p{1.6cm}
      >{\centering\arraybackslash}p{1.6cm}
      >{\centering\arraybackslash}p{1.6cm}
      >{\columncolor{gray!15}\centering\arraybackslash}p{1.5cm}
      | >{\centering\arraybackslash}p{1.5cm}}
        \toprule
  \textbf{Model} &
    \textbf{Fluency $\uparrow$} & \textbf{Dyadic Affect Match $\uparrow$} & \textbf{Nonverbal Cue Approp. $\uparrow$} & \textbf{Overall $\uparrow$} & \textbf{Timing $\uparrow$} \\
   \midrule
 Human ground truth & 4.42 {\scriptsize (4.21--4.61)} & 4.14 {\scriptsize (3.86--4.40)} & 3.18 {\scriptsize (2.78--3.59)} & 3.92 & \makecell{78\% \\ {\scriptsize 900\,ms}} \\
   \midrule
Gemini 2.5 + Anam~\cite{anam2026} & \textbf{3.48} {\scriptsize (3.17--3.79)} & \textbf{3.21} {\scriptsize (2.88--3.55)} & \textbf{1.71} {\scriptsize (1.27--2.15)} & \textbf{2.80} & \makecell{\textbf{44\%} \\ {\scriptsize \textbf{2840\,ms}}} \\
 Gemini 2.5 + Keyframe~\cite{keyframe2026} & 3.43 {\scriptsize (3.12--3.74)} & 2.60 {\scriptsize (2.28--2.93)} & 1.13 {\scriptsize (0.75--1.50)} & 2.39 & \makecell{31\% \\ {\scriptsize 3520\,ms}} \\

   \bottomrule
 \end{tabular}
 }
 \vskip -1mm
\end{table*}
 
\subsection{Vision-perceiving speech models}
\label{sec:results-Vision-perceiving}

\textbf{\textit{Insight 1}. Current models remain below human-level conversational naturalness.}
\Cref{tab:leaderboard-perception} shows no model approaches human ground truth on \name{}; the largest deficits concentrate in fast social-coordination dynamics (Pause Handling, Nonverbal Backchanneling, Gaze Avoidance with Pause), with aggregate human--model gaps up to $0.85$ (\Cref{tab:per-dynamic-results-perception}). Open-source MiniCPM-o-4.5~\cite{minicpmo2025} leads in both AV and AO overall (3.40/3.44), with Gemini 2.5 Flash Native as runner-up. Conversational Flow shows the largest gap: humans score 4.20, closed-source AV2A systems score 2.20--2.81, and the strongest AV2A model reaches only 3.54. TOR-Alignment follows the same pattern: humans score $90\%$/$1400$\,ms vs.\ next-best MiniCPM-o~4.5 at $73\%$/$720$\,ms.

\textbf{\textit{Insight 2}. Limited visual frame rate prevents models from capturing nonverbal conversational signals.}
Audio is processed at millisecond-scale temporal resolution, but AV2A models typically take visual input at 1 FPS, missing nonverbal dynamics that unfold within $1$--$2$ seconds.
Taking Nonverbal Interruption as an example, agents are expected to yield within $1.5$\,s, but Gemini 3.1 achieves only TOR-Alignment of $40\%$/$60\%$ (AV2A/A2A) with $1720$/$1240$\,ms median latency; Gemini 2.5 only scores $20\%$/$20\%$ on TOR-Alignment and even higher latency.

MiniCPM-o~4.5 uniquely exposes a user-controllable visual sampling rate knob, so we explore the impact of video sampling rates $\{1\text{--}10\}$ FPS.\footnote{We selected this sweep to bridge the model's reported FPS configurations: public documentation recommends inference at 5--10 FPS, while their paper describes training with 1--5 FPS.}
We find that performance peaks at 2 FPS, then declines as FPS increases (\name{} overall: $3.04$ at 8 FPS, $2.81$ at 10 FPS; fluency: $3.55 \rightarrow 2.33$); more concerningly, output speech becomes increasingly incoherent.
This pattern suggests a vision--speech fusion bottleneck: denser visual input can overload the shared cross-modal attention budget and degrade response quality.

\textbf{\textit{Insight 3}. No system successfully leverages the visual channel for both timing and content.}
Comparing audio-only speech agents with audio-video speech agents (\cref{tab:leaderboard-perception}) reveals large variance in scores across model families.
In all cases, the audio-video speech agents score worse on \name{}'s perception rubrics than the audio-only speech agents.
We observe three distinct patterns:

\begin{wrapfigure}{r}{0.48\textwidth}
  \centering
  \vskip -4mm
  \includegraphics[width=\linewidth]{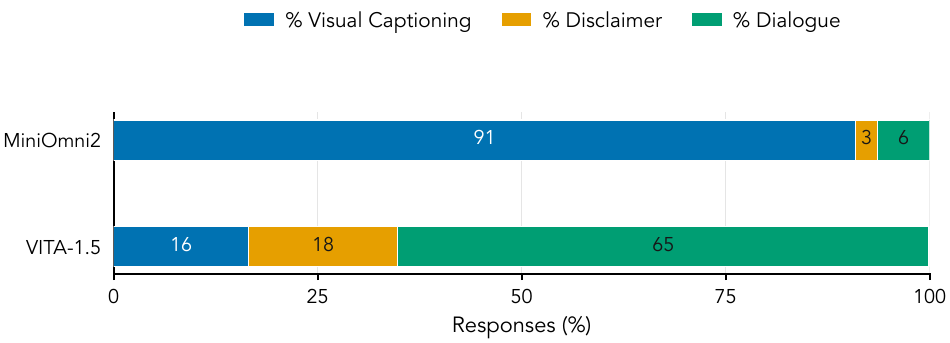}
  \vskip -2mm
  \caption{Mini-Omni2~\cite{xie2024miniomni2} and VITA-1.5~\cite{fu2026vita} often respond to vision-speech input with visual captions or blanket disclaimers.}
  \vskip -4mm
  \label{fig:captioning-mode-failure}
\end{wrapfigure}

\begin{itemize}[leftmargin=*]
\setlength\parskip{0em}
\item \textbf{Captioning collapse.}         
We find that many AV models treat visual inputs as captioning prompts in \name{}. 
To quantify this, we classify responses as \{\texttt{dialogue}, \texttt{visual captioning}, \texttt{disclaimers}\} with an LM judge (\cref{fig:captioning-mode-failure}). 
Mini-Omni2 responds with \texttt{visual captioning} on 87\% of clips, but reverts to dialogue in audio-only mode. 
VITA-1.5 shows a lower captioning rate (17\%) but exhibits other failures, such as blanket capability disclaimers (``\textit{I'm just a computer program and I don't have the ability to see or hear things\ldots}'') and token doubling in \(\sim 74\%\) of responses.
A milder version of this pattern appears in \texttt{gpt-realtime}: on AV perception clips, the full model intermittently starts with visual-scene framing (e.g., ``I can see you're \ldots'', ``you look \ldots'', ``it looks like \ldots'').

\item \textbf{Visual-stream ignorance.}
\texttt{gpt-realtime-mini} manifests a different consequence: it produces AV2A and A2A outputs that are paraphrases of each other; visual inspection of side-by-side transcripts confirms the visual stream rarely shifts response timing or content. This indicates the visual stream is not leveraged to provide any additional context or information.

\item \textbf{TOR-Alignment.} 
Timing in \Cref{tab:leaderboard-perception} shows a consistent pattern across models: relative to audio-only, visual input reduces TOR-Alignment by $0$--$5$ points for all six models except OpenAI \texttt{gpt-realtime}. Although \texttt{gpt-realtime} gains $+5$ points in TOR-Alignment, its median latency increases $1000$\,ms, so the response remains slower overall. \emph{Adding video does not improve timing while staying within realtime tolerance for any model.} 
MiniCPM-o-4.5 is evaluated locally and shows strong timing in both modes ($73\%$ / $720$\,ms AV; $72\%$ / $920$\,ms AO). 
\end{itemize}

We find that no model achieves simultaneous AV2A gains over its audio-only counterpart on all perception axes. 
Our results indicate current systems primarily use visual input for explicit visual question answering rather than for robust joint audiovisual grounding in natural conversation. 
As a consequence, A2A models actually outperform their AV2A counterparts on \name{}'s natural audiovisual dialogue.

\subsection{Cascaded-avatar speech models}
\label{sec:results-Vision-generation}

\textbf{\textit{Insight 4}. Cascading a full-duplex speech model with an avatar-rendering layer preserves turn-taking discipline but cannot supply real-time nonverbal cue production.}
Relative to human ground truth, cascaded avatars show only a modest drop in Fluency ($4.42 \rightarrow 3.43$--$3.48$) but a large drop in Nonverbal Cue Appropriateness ($3.18 \rightarrow 1.13$--$1.71$). This gap is expected because audio-driven avatars are effectively turn-based (motion follows produced speech only), so they cannot add cues during the user's turn, and Anam/Keyframe cascade latency ($2840$--$3520$\,ms) is too high for interactive nonverbal timing. Overall, cascaded A2AV pipelines remain fundamentally limited for nonverbal communication, motivating end-to-end speech-vision models or avatar layers that emit nonverbal motion independently of speech.

\section{Conclusion}
\label{sec:conclusion}

We introduced \name{}, the first benchmark for evaluating audio-visual full-duplex conversational agents, with \numdynamics{} nonverbal dynamics, \numsamples{} expert-annotated clips, and rubric-based LM-as-judge scoring for perception and generation.
Our analysis finds that current systems remain well below human conversational naturalness: AV2A models do not improve over A2A baselines and often fail to integrate nonverbal signals (e.g., \emph{captioning collapse} and \emph{visual-stream ignorance}), while cascaded A2AV avatars cannot produce independent, real-time nonverbal cues.
These findings highlight the need for tighter joint audio-visual grounding in full-duplex interaction.
We will release benchmark data and evaluation code to accelerate progress toward full-duplex audio-visual agents that converse naturally with humans.

\textbf{Limitations \& Future Work.} \name{} has three main limitations: (1) our dataset is limited to English-only conversations, (2) our mid-conversation system prompts are not always strong enough to override a model's pretrained greeting defaults, and (3) \name{} depends on LM-judge quality which, in turn, depends on upstream caption fidelity. Future work should expand behavior coverage (full-body cues, multi-turn interactions) and broaden data to embodied settings.

\textbf{Impacts.} \name{} benchmarks progress toward full-duplex multimodal conversational agents.
As humanistic conversational AI systems expand into high-impact settings, both misuse and miscommunication can cause harm.
We advocate for rigorous pre-deployment evaluation and explicit safety guardrails. 
\begin{ack}
We thank David Luebke, Ruth Rosenholtz, Ekta Prashnani, Slim Essid, and Viet Anh Trinh for feedback and early discussions on the project. 
\end{ack}

{
\small
\bibliographystyle{plainnat}
\bibliography{main}
}

\appendix

\newpage
\section*{Appendix}

\section{Dataset}

\subsection{Additional Dataset Details}
\paragraph{Technical specs.}
All source recordings are screened to meet minimum quality thresholds (Table~\ref{tab:tech-specs}).

\begin{table}[h]
  \centering
  \caption{Minimum technical specifications for VideoFDB source recordings.}
  \label{tab:tech-specs}
  \small
  \setlength{\tabcolsep}{10pt}
  \begin{tabular}{ll}
    \toprule
    \textbf{Property} & \textbf{Minimum} \\
    \midrule
    Video resolution & 720p \\
    Frame rate & 30 fps \\
    Audio sample rate & 24 kHz \\
    Video format & VP9 (\texttt{.mp4}) \\
    Audio format & PCM (\texttt{.wav}) \\
    \bottomrule
  \end{tabular}
\end{table}

\paragraph{Collection and annotation details.}
\label{sec:appendix-dataset-collection}
Our dataset is mined from a large corpus of two-person video calls, and specifically held out from other dataset releases to prevent training data contamination.
Participants are connected via a videoconferencing platform and given a conversation prompt, which one interlocutor introduces naturally to initiate the conversation.
To minimize the effects of network latency, each speaker's audio-video stream is recorded locally in parallel with live transmission to the other participant.

Source recordings are captured locally and passed through media-quality and preprocessing checks, including intra-channel A/V sync validation and inter-channel alignment validation.
Dataset construction follows a three-pass annotation workflow: (1) human annotators manually identify candidate dynamic moments in longer recordings, (2) additional human annotators validate the trimmed clips, retaining only candidate samples with sufficient pre-event context (targeting 1--3 turns prior to the event), and assign precise event timestamps and dynamic type labels, and (3) a final human reviewer performs quality checks for label consistency and timestamp precision.
Validated clips are then passed through duration verification with \texttt{ffprobe}, category filtering to in-scope dynamics, and trim validation before release packaging.

\paragraph{Summary statistics.}
Table~\ref{tab:dataset-stats} summarizes clip and dynamic-window durations for the full set and the curated evaluation subset.

 \begin{table}[h]
    \centering
    \caption{Dataset summary statistics. Dynamic-window length is $t_\text{end}-t_\text{start}$. }
    \label{tab:dataset-stats}
    \small
    \setlength{\tabcolsep}{7pt}
    \resizebox{\textwidth}{!}{%
    \begin{tabular}{lrrrr}
      \toprule
      \textbf{Set} & \textbf{\# clips} & \textbf{\# P / \# G} & \textbf{Clip duration (median / IQR)} & \textbf{Dyn.\
  window (median / IQR)} \\
      \midrule
      Test (held-out scoring set)  & 226 & 105 / 121 & 46.0s / [34.0, 61.4] & 2.5s / [2.0, 5.0] \\
      Validation (public eval set) & 11  & 5 / 6     & 46.0s / [37.6, 55.8] & 6.0s / [2.0, 28.2] \\
      \bottomrule
    \end{tabular}%
    }
  \end{table}

\paragraph{Demographic statistics.}
Our dataset includes English-speaking participants with a range of accents and backgrounds, located in the United States and Canada. Table~\ref{tab:demographic-stats} summarizes the demographic statistics of the dataset.
\begin{table}[h]
  \centering
  \caption{Demographic statistics of the dataset.}
  \label{tab:demographic-stats}
  \small
  \setlength{\tabcolsep}{7pt}
  \begin{tabular}{ll}
    \toprule
    \textbf{Property} & \textbf{Value} \\
    \midrule
    Unique speakers & 130 \\
    \midrule
    \multicolumn{2}{l}{\textbf{Age Range (Percent of Speakers)}} \\
    18--29 & 19\% \\
    30--39 & 32\% \\
    40--49 & 20\% \\
    50--59 & 18\% \\
    60--69 & 8\% \\
    70--79 & 2\% \\
    \midrule
    \multicolumn{2}{l}{\textbf{Gender (Percent of Speakers)}} \\
    Female & 44\% \\
    Male & 54\% \\
    Prefer Not to Say & 2\% \\
    \bottomrule
  \end{tabular}
\end{table}

\section{Further Limitations}
\paragraph{Dataset scope and representational constraints.} The benchmark contains English-language, two-person video conferencing conversations recorded by participants in the United States and Canada, representing a narrow slice of video-call interaction. All recordings consist of dyadic webcam-style captures from standard video conferencing setups; they do not represent in-person, mobile, or alternative camera configurations. 
Cultural context further modulates nonverbal communication channels, since gesture, gaze, and affect conventions vary across communities and interaction settings.
\name{} is therefore not recommended for evaluation of multilingual models, in-person or multi-party conversation modeling, or non-English-speaking populations.
\paragraph{Evaluation scope.} The benchmark supports single-turn evaluation of audiovisual conversational dynamics only. It does not support multi-turn evaluation, training, or fine-tuning of any kind, nor generalization to codec environments, recording modalities, or conversational contexts outside those represented here.
\paragraph{Evaluation pipeline constraints.} Like all LM-based evaluations, our evaluation is bounded by the perceptual capabilities of the underlying captioning model, including both its visual and audio comprehension quality. Specifically, we observe that our judge assessment of ground-truth humans defines an upper bound for rubric scoring.

\section{Evaluation Protocol Details}
\label{sec:appendix-eval-protocol}

\subsection{Dynamic-to-rubric mapping}
Table~\ref{tab:dynamic-axis-map} shows the full breakdown of which rubric axes apply to each conversational dynamic, split between perception and generation buckets.

\begin{table*}[t]
  \centering
  \caption{A conversational dynamic can be evaluated under perception, generation, or both. ``\#'' reports total samples per dynamic across perception and generation splits.}
  \label{tab:dynamic-axis-map}
  \small
  \setlength{\tabcolsep}{5pt}
  \resizebox{\textwidth}{!}{
  \begin{tabular}{lccccccc}
    \toprule
    \textbf{Dynamic} & \textbf{\#} & \multicolumn{3}{c}{\textbf{Perception}} & \multicolumn{3}{c}{\textbf{Generation}} \\
    \cmidrule(lr){3-5} \cmidrule(lr){6-8}
    &  & \textbf{Fluency} & \textbf{Conv. Flow} & \textbf{Semantic Grounding} & \textbf{Fluency} & \textbf{Dyadic Affect Match} & \textbf{Cue. Approp.} \\
    \midrule
    Pause Handling & 13 & \cmark & \cmark &  &  &  &  \\
    Gaze Avoidance with Pause & 18 & \cmark & \cmark &  &  &  &  \\
    Nonverbal Interruption & 8 & \cmark & \cmark & \cmark &  &  &  \\
    Adaptor Handling & 14 & \cmark &  & \cmark &  &  &  \\
    Face Emotion Display & 50 & \cmark &  & \cmark & \cmark & \cmark & \cmark \\
    Laughter & 30 & \cmark &  & \cmark & \cmark & \cmark & \cmark \\
    Nonverbal Backchanneling & 34 & \cmark & \cmark &  & \cmark & \cmark & \cmark \\
    Emotion Matching & 13 &  &  &  & \cmark & \cmark & \cmark \\
    Verbal Backchanneling & 17 &  &  &  & \cmark & \cmark & \cmark \\
    Verbal Interruption & 24 &  &  &  & \cmark & \cmark & \cmark \\
    Turn-taking & 16 &  &  &  & \cmark & \cmark &  \\
    \bottomrule
  \end{tabular}
  }
\end{table*}

\subsection{Timing Metrics Details}\label{sec:appendix-timing-metrics}
TOR-Alignment extends the speech-only TOR formulation of \cite{lin2025fdb_v1} to multimodal AV dynamics and places heterogeneous timing expectations under one metric.
Rather than inferring turn boundaries from ASR alone, we evaluate timing against the annotated event window \texttt{[dynamic\_start\_s, dynamic\_end\_s]}.
We compute TOR-Alignment for each timing-relevant dynamic by mapping it to one of five \textbf{timing classes}, defined by the expected agent behavior at the cue (see Tab.~\ref{tab:dynamic-axis-map}):
\begin{itemize}
\itemsep0pt 
\item \textsc{stay-silent} (Pause Handling, Gaze Avoidance with Pause): the user is mid-thought; the agent should not take the floor in-window. A clip passes if in-window agent speech is $\leq$\,1\,s, or all overlap segments are backchannels ($<$\,1\,s and $<$\,2 words; \cite{lin2025fdb_v1}).
\item \textsc{continue-speaking} (Nonverbal Backchanneling, Adaptor Handling): the user emits a non-floor-offering cue while the agent is speaking; failure is inappropriate yielding. If pre-cue agent activity in the previous 3\,s is $\geq$\,30\%, in-window activity must remain $\geq$\,50\%; otherwise, the case is treated as vacuously correct.
\item \textsc{yield-required} (Nonverbal Interruption, Verbal Interruption generation): the user takes the floor; the agent must yield within 1.5\,s. Latency is measured as the end time of the agent segment overlapping (or immediately preceding) \texttt{dynamic\_start\_s}, relative to \texttt{dynamic\_start\_s}.
\item \textsc{smooth-handoff} (Turn-taking): the agent should start speaking within the annotated handoff window. Latency is \texttt{first\_onset\_after(dynamic\_start\_s)}; if the agent is already speaking at cue onset, latency is 0\,ms.
\item \textsc{backchannel-produced} (Verbal Backchanneling): the agent should produce at least one in-window backchannel ($<$\,1\,s, $<$\,2 words) and avoid full floor-taking (no segment above backchannel limits). Multiple short in-window backchannels are valid.
\end{itemize}

\paragraph{Formal definition.}
Following \cite{lin2025fdb_v1}, the binary takeover variable is
\begin{equation*}
\mathrm{TO}_i = \begin{cases} 0, & \text{if the agent's output is silence or a backchannel,} \\ 1, & \text{otherwise,} \end{cases}
\end{equation*}
for each clip $i$, and the takeover rate is $\mathrm{TOR} = \frac{1}{N}\sum_{i=1}^{N} \mathrm{TO}_i$ -- typically reported per dynamic, with the preferred direction (lower-better or higher-better) interpreted case-by-case.
In TOR-Alignment, we encode that direction directly: each timing class $c \in \mathcal{C}$ has a policy-prescribed expected takeover $\mathrm{TO}^*_c \in \{0, 1\}$ --- $0$ for \textsc{stay-silent}, \textsc{yield-required}, and \textsc{backchannel-produced}; $1$ for \textsc{continue-speaking} and \textsc{smooth-handoff}.
The per-clip alignment indicator is
\begin{equation*}
A_i \;=\; \begin{cases} 1, & \text{if } \mathrm{TO}_i = \mathrm{TO}^*_{c_i}, \\ 0, & \text{otherwise,} \end{cases}
\end{equation*}
and TOR-Alignment is its mean:
\begin{equation*}
\mathrm{TOR\text{-}Alignment} \;=\; \frac{1}{N} \sum_{i=1}^{N} A_i.
\end{equation*}

Higher TOR-Alignment indicates better agreement with timing-class-specific takeover policy, i.e., the fraction of clips that satisfy the expected timing behavior.

\subsection{Judge Validation}
\label{sec:appendix-judge-agreement}

We assess judge robustness across three judge backends: \texttt{meta/llama-3.1-70b-instruct}, \texttt{azure/openai/gpt-4o}, and \texttt{azure/anthropic/claude-sonnet-4-6}. Each judge independently scores the same response on each axis with the same rubric and prompt. We report (i) ground truth (GT) calibration on the held-out reference set (Table~\ref{tab:judge-model-comparison}) and (ii) inter-judge agreement on a combined GT+RANDOM cross-judge subset of $n{=}220$ clips ($n{=}110$ ground-truth + $n{=}110$ random-baseline with mixed Initiator/Respondent channels; Table~\ref{tab:judge-agreement}).

\paragraph{GT calibration.}
Table~\ref{tab:judge-model-comparison} reports judge-specific mean$\pm$stdev across Fluency, Conversational Flow, and Visual Grounding, plus a consensus estimate with 95\% confidence intervals. The three judges are closely aligned on aggregate GT means while still exposing judge-dependent spread (stdev), motivating reporting both per-judge variability and consensus confidence intervals.

\begin{table*}[t]
  \centering
  \caption{GT judge calibration across three judge backends. Judge-specific cells report mean $\pm$ stdev (0--5). The final row reports consensus statistics (per-sample mean across the three judges) with 95\% CI and sample count.}
  \label{tab:judge-model-comparison}
  \small
  \setlength{\tabcolsep}{6pt}
  \resizebox{\textwidth}{!}{%
  \begin{tabular}{lccc}
    \toprule
    \textbf{Judge} & \textbf{Fluency} & \textbf{Conversational Flow} & \textbf{Visual Grounding} \\
    \midrule
    \texttt{llama-3.1-70b-instruct} & $4.52 \pm 0.94$ & $4.50 \pm 0.91$ & $4.26 \pm 0.54$ \\
    \texttt{gpt-4o}               & $4.54 \pm 1.28$ & $4.48 \pm 1.10$ & $4.18 \pm 1.06$ \\
    \texttt{claude-sonnet-4-6} & $4.55 \pm 0.79$ & $4.16 \pm 0.95$ & $4.56 \pm 0.59$ \\
    \midrule
    \textbf{Consensus (3-judge mean)}    & \textbf{4.53 (95\% CI: [4.38, 4.69]} & \textbf{4.38 (95\% CI: [4.16, 4.60] )} & \textbf{4.33 (95\% CI: [4.18, 4.48] } \\
    \bottomrule
  \end{tabular}%
  }
\end{table*}

\paragraph{Inter-judge agreement.}
We find $77$--$89\%$ pairwise agreement within $1$ point on the $0$--$5$ scale: the 3-judge averaged score reaches $\mathrm{ICC(A,k)} = 0.84$ / $0.90$ / $0.75$ on perception rubrics Fluency / Conversational Flow / Visual Grounding (Table~\ref{tab:judge-agreement}). We use the intraclass correlation coefficient (ICC) under a two-way random-effects model with absolute-agreement criterion. ICC(A,1) is the reliability of a single judge's score; ICC(A,k) is the reliability of the averaged score across the $k{=}3$ judges --- the relevant statistic for the consensus mean reported on the leaderboard. We follow the \citet{Koo2016AGO} interpretation: $<\!0.50$ poor, $0.50$--$0.75$ moderate, $0.75$--$0.90$ good, $>\!0.90$ excellent.

\begin{table}[t]
  \centering
  \caption{Inter-judge agreement on perception rubrics across three judge backends (\texttt{llama-70b}, \texttt{gpt-4o}, \texttt{claude-sonnet-4.6}) on the combined GT+RANDOM subset. We report ICC (single-judge and averaged-judge), Within-1pt agreement on the 0--5 scale, and MAD (mean absolute pairwise difference).}
  \label{tab:judge-agreement}
  \small
  \begin{tabular}{l c c c c c}
    \toprule
    Axis & $n$ & ICC(A,k) [95\% CI] & ICC(A,1) [95\% CI] & Within-1pt & MAD \\
    \midrule
    Fluency             & 220 & $0.84$ \scriptsize[$0.79$, $0.87$] & $0.63$ \scriptsize[$0.55$, $0.70$] & $77.7\%$ & $0.98$ \\
    Conversational Flow & 112 & $0.90$ \scriptsize[$0.86$, $0.93$] & $0.76$ \scriptsize[$0.67$, $0.83$] & $88.7\%$ & $0.70$ \\
    Visual Grounding    & 124 & $0.75$ \scriptsize[$0.66$, $0.82$] & $0.50$ \scriptsize[$0.39$, $0.60$] & $80.6\%$ & $0.82$ \\
    \bottomrule
  \end{tabular}
\end{table}

The 3-judge averaged score reaches good-to-excellent reliability for Fluency and Conversational Flow ($\mathrm{ICC(A,k)} \geq 0.84$), and moderate-to-good reliability for Visual Grounding ($\mathrm{ICC(A,k)} = 0.75$, lower 95\% CI bound $0.66$). At the single-judge level, agreement is moderate for Fluency and Conversational Flow ($\mathrm{ICC(A,1)} = 0.63$ and $0.76$) and weaker for Visual Grounding ($\mathrm{ICC(A,1)} = 0.50$).

This pattern is consistent with the nature of the task. Fluency and Conversational Flow can be reasonably assessed from the response transcript and turn-state metadata, whereas Visual Grounding requires judges to determine whether the response is conditioned on a visual cue described in an auxiliary caption; this judgment has greater scoring variability. Even on this most challenging axis, $80.6\%$ of pairwise (clip, axis) comparisons are within one point on the 0--5 scale, and averaging across three judges raises reliability to the moderate-to-good range. We therefore interpret cross-model differences in Visual Grounding at the model-aggregate level over the $n{=}105$ test clips, where the standard error of the mean is substantially smaller than per-clip disagreement.

\section{Models \& Implementation}

We evaluate a diverse set of full-duplex baselines spanning omni-modal, audio-visual, and audio-only conversational speech agents.

\paragraph{System prompts.} Prompt-construction details are provided in \cref{sec:judge-prompts-system-prompt}; the only model tested that does not support system prompts is MiniOmni2~\cite{xie2024miniomni2}.

\paragraph{Model details.} We evaluate the following models with the same
\name{} protocol: each clip is streamed through a unified realtime
harness, using audio+video for AV runs and disabling video for audio-only
runs. Unless noted otherwise, we provide the clip-specific \name{} system
prompt to the model.
\begin{itemize}
\item \textbf{Gemini 2.5 / 3.1 Flash Native}~\cite{comanici2025gemini}. We
run Gemini models with the Gemini API and WebSocket input, streaming user audio and sampled
video frames. We then log the model's spoken response for scoring.

\item \textbf{OpenAI Realtime / Realtime mini}~\cite{openai_realtime_api}.
We run OpenAI Realtime with the OpenAI API and WebSocket input, with the same clip-level streaming setup and prompt
construction used for Gemini so comparisons stay controlled.

\item \textbf{MiniCPM-o 4.5}~\cite{minicpmo2025}. We run
MiniCPM-o-4.5 directly through its full-duplex demo API. Audio is streamed in 1-second chunks; for
AV runs we provide sampled video frames, and for audio-only runs we disable
frame input. We also use one fixed reference TTS clip across samples for
consistency.

\item \textbf{MiniOmni2}~\cite{xie2024miniomni2}. We integrate MiniOmni2
directly, using its AV path when frames are
available and its audio-only path otherwise. Because its serving interface
is ultimately half-duplex, we buffer each clip, segment speech with Silero VAD~\citep{silero_vad}, align each segment to the most recent preceding
video frame, and merge segment-level outputs into one response. To fit its
fixed Whisper input length, we cap each segment at 29.5 seconds and warn
when truncation occurs. MiniOmni2 does not support system prompts.

\item \textbf{VITA-1.5}~\cite{fu2026vita}. We deploy VITA-1.5 using their
official realtime demo stack~\citep{fu2026vita}. We replace the default prompt (``I am an AI robot named
VITA'') with ``You are a helpful assistant on a video call.'' and append
our clip-specific system prompt. We evaluate both AV
($n_{\text{frames}}=4$ at 1 fps) and audio-only (video disabled). For clips
marked \texttt{agent\_speaks\_first}, we trigger an initial no-audio turn
after the first video frame.

\item \textbf{Gemini Live 2.5 Flash Native + Anam avatar}~\cite{google_gemini_live_api, anam2026}.
This is a cascaded setup: Gemini generates the speech response, and Anam
renders avatar motion from that speech stream.

\item \textbf{Gemini Live 2.5 Flash Native + Keyframe avatar}~\cite{google_gemini_live_api, keyframe2026}.
This matches the same cascaded pipeline as above, but uses Keyframe as the
avatar renderer.
\end{itemize}

\paragraph{Hardware Details}
For open-source models, we run local inference on a server with 8$\times$NVIDIA H100 80GB HBM3 GPUs and 2$\times$Intel(R) Xeon(R) Platinum 8480+ CPUs, using one GPU per run.

\section{Per-Dynamic Results}
\label{sec:per-dynamic-results}

\Cref{tab:per-dynamic-results-perception,tab:per-dynamic-results-generation,tab:per-dynamic-results-generation-timing} present per-category call-outs split into Perception, Generation rubrics, and Generation timing.

\begin{table}[t]
  \centering
  \caption{Per-category perception results.}
  \label{tab:per-dynamic-results-perception}
  \small
  \renewcommand{\arraystretch}{1.15}
  \resizebox{\columnwidth}{!}{
  \begin{tabular}{l | c c c c c c c}
    \toprule
    \textbf{Model} &
      \textbf{Face Emotion} & \textbf{Gaze Avoid.} & \textbf{Nonverb. BC} & \textbf{Laughter} & \textbf{Adaptor} & \textbf{Pause} & \textbf{Nonverb. Int.} \\
    \midrule
    Human reference & 4.08 {\scriptsize (3.62--4.46)} & 3.71 {\scriptsize (2.76--4.53)} & 4.12 {\scriptsize (4.00--4.31)} & 3.86 {\scriptsize (3.14--4.50)} & 5.00 {\scriptsize (5.00--5.00)} & 5.00 {\scriptsize (5.00--5.00)} & 4.12 {\scriptsize (3.81--4.44)} \\
    \midrule
    \multicolumn{8}{l}{\textit{Closed-source full-duplex speech-vision models}} \\
    \midrule
    Gemini direct (AV) & 2.42 {\scriptsize (1.75--3.08)} & 2.94 {\scriptsize (1.76--4.12)} & 2.44 {\scriptsize (1.50--3.56)} & 3.57 {\scriptsize (3.00--4.00)} & 5.00 {\scriptsize (5.00--5.00)} & 3.46 {\scriptsize (1.92--4.62)} & 2.75 {\scriptsize (1.69--3.62)} \\
    Gemini 3.1 direct (AV) & 2.61 {\scriptsize (1.92--3.29)} & 2.00 {\scriptsize (1.25--2.81)} & 2.85 {\scriptsize (2.06--3.65)} & 3.73 {\scriptsize (3.27--4.14)} & 4.90 {\scriptsize (4.70--5.00)} & 2.73 {\scriptsize (1.77--3.65)} & 2.33 {\scriptsize (1.67--3.00)} \\
    OpenAI gpt-realtime-mini (AV) & 1.83 {\scriptsize (1.08--2.50)} & 2.94 {\scriptsize (1.76--4.12)} & 2.19 {\scriptsize (1.12--3.31)} & 3.00 {\scriptsize (2.43--3.50)} & 5.00 {\scriptsize (5.00--5.00)} & 2.85 {\scriptsize (1.62--4.08)} & 1.62 {\scriptsize (0.81--2.44)} \\
    OpenAI gpt-realtime (AV) & 2.12 {\scriptsize (1.33--2.92)} & 2.88 {\scriptsize (1.76--4.06)} & 2.62 {\scriptsize (1.62--3.62)} & 3.21 {\scriptsize (2.57--3.79)} & 4.62 {\scriptsize (3.85--5.00)} & 2.31 {\scriptsize (1.15--3.85)} & 2.25 {\scriptsize (1.31--3.38)} \\
    \midrule
    \multicolumn{8}{l}{\textit{Open-source full-duplex speech-vision models}} \\
    \midrule
    MiniCPM-o 4.5 & 3.00 {\scriptsize (2.33--3.58)} & 3.24 {\scriptsize (2.12--4.35)} & 3.44 {\scriptsize (2.56--4.19)} & 3.07 {\scriptsize (2.29--3.79)} & 5.00 {\scriptsize (5.00--5.00)} & 3.85 {\scriptsize (2.69--5.00)} & 4.06 {\scriptsize (3.75--4.38)} \\
    MiniOmni2 (AV) & 0.83 {\scriptsize (0.58--1.08)} & 0.59 {\scriptsize (0.00--1.47)} & 3.00 {\scriptsize (2.00--3.75)} & 0.50 {\scriptsize (0.21--0.86)} & 3.85 {\scriptsize (2.69--5.00)} & 0.38 {\scriptsize (0.00--1.15)} & 1.56 {\scriptsize (0.62--2.75)} \\
    VITA-1.5 (AV) & 2.08 {\scriptsize (1.62--2.58)} & 0.65 {\scriptsize (0.00--1.53)} & 2.75 {\scriptsize (1.75--3.50)} & 1.21 {\scriptsize (0.71--1.71)} & 4.15 {\scriptsize (3.08--5.00)} & 0.00 {\scriptsize (0.00--0.00)} & 3.62 {\scriptsize (3.25--4.00)} \\
    \midrule
    \multicolumn{8}{l}{\textit{Audio-only full-duplex speech-vision models}} \\
    \midrule
    VITA-1.5 (audio only) & 2.38 {\scriptsize (1.83--2.92)} & 0.00 {\scriptsize (0.00--0.00)} & 2.94 {\scriptsize (2.06--3.75)} & 2.21 {\scriptsize (1.50--3.00)} & 5.00 {\scriptsize (5.00--5.00)} & 0.00 {\scriptsize (0.00--0.00)} & 3.25 {\scriptsize (2.31--4.00)} \\
    Gemini direct (audio only) & 2.04 {\scriptsize (1.29--2.83)} & 2.94 {\scriptsize (1.76--4.12)} & 2.81 {\scriptsize (1.75--3.94)} & 3.21 {\scriptsize (2.71--3.71)} & 5.00 {\scriptsize (5.00--5.00)} & 3.85 {\scriptsize (2.69--5.00)} & 2.75 {\scriptsize (1.81--3.56)} \\
    Gemini 3.1 direct (audio only) & 2.63 {\scriptsize (1.97--3.29)} & 2.19 {\scriptsize (1.42--3.00)} & 3.76 {\scriptsize (3.12--4.35)} & 3.91 {\scriptsize (3.50--4.32)} & 4.25 {\scriptsize (3.45--4.95)} & 3.08 {\scriptsize (2.12--4.04)} & 2.76 {\scriptsize (2.14--3.38)} \\
    OpenAI gpt-realtime-mini (audio only) & 2.00 {\scriptsize (1.33--2.67)} & 2.65 {\scriptsize (1.53--3.82)} & 2.69 {\scriptsize (1.69--3.75)} & 3.07 {\scriptsize (2.57--3.57)} & 5.00 {\scriptsize (5.00--5.00)} & 2.31 {\scriptsize (1.15--3.85)} & 2.75 {\scriptsize (2.00--3.38)} \\
    OpenAI gpt-realtime (audio only) & 3.04 {\scriptsize (2.25--3.79)} & 2.41 {\scriptsize (1.29--3.59)} & 3.12 {\scriptsize (2.12--4.19)} & 3.43 {\scriptsize (2.79--4.00)} & 5.00 {\scriptsize (5.00--5.00)} & 1.54 {\scriptsize (0.38--3.08)} & 2.69 {\scriptsize (1.69--3.56)} \\
    MiniCPM-o 4.5 (audio only) & 1.67 {\scriptsize (0.92--2.46)} & 3.76 {\scriptsize (2.71--4.65)} & 3.31 {\scriptsize (2.50--4.06)} & 3.50 {\scriptsize (2.86--4.07)} & 5.00 {\scriptsize (5.00--5.00)} & 4.62 {\scriptsize (3.85--5.00)} & 3.44 {\scriptsize (2.94--3.81)} \\
    MiniOmni2 (audio only) & 0.94 {\scriptsize (0.62--1.27)} & 1.59 {\scriptsize (0.85--2.35)} & 1.44 {\scriptsize (0.84--2.06)} & 1.61 {\scriptsize (1.14--2.11)} & 3.58 {\scriptsize (2.85--4.23)} & 1.54 {\scriptsize (0.77--2.50)} & 2.12 {\scriptsize (1.38--2.92)} \\
    \bottomrule
  \end{tabular}
  }
\end{table}

\begin{table}[t]
  \centering
  \caption{Per-category generation-axis results (means). \textit{Hum.} is the hand-annotated human reference; \textit{Anam} and \textit{KF} are Gemini 2.5 Flash Native cascaded with the Anam (neural rendering) and Keyframe (interpolation) avatar layers respectively. }
  \label{tab:per-dynamic-results-generation}
  \small
  \renewcommand{\arraystretch}{1.15}
  \resizebox{\columnwidth}{!}{
  \begin{tabular}{l | c c c | c c c | c c c}
    \toprule
    & \multicolumn{3}{c|}{\textbf{Gen.\ Fluency $\uparrow$}} & \multicolumn{3}{c|}{\textbf{Affect Match $\uparrow$}} & \multicolumn{3}{c}{\textbf{Cue Appropr.\ $\uparrow$}} \\
    \cmidrule(lr){2-4} \cmidrule(lr){5-7} \cmidrule(lr){8-10}
    \textbf{Category} & Hum. & Anam & KF & Hum. & Anam & KF & Hum. & Anam & KF \\
    \midrule
    Laughter                  & 4.14 & 2.86 & \textbf{3.07} & 4.14 & \textbf{2.00} & 1.14 & 3.57 & \textbf{0.14} & 0.00 \\
    Nonverbal Backchanneling  & 4.38 & 3.25 & \textbf{3.62} & 4.44 & \textbf{4.25} & 4.00 & 1.75 & \textbf{0.00} & \textbf{0.00} \\
    Verbal Backchanneling     & 4.19 & 3.12 & \textbf{3.19} & 4.56 & \textbf{2.81} & 2.25 & 3.31 & \textbf{0.06} & 0.00 \\
    Verbal Interruption       & 4.13 & \textbf{3.43} & 3.00 & 2.74 & \textbf{3.96} & 3.09 & 2.27 & \textbf{3.12} & 1.36 \\
    Emotion Matching          & 4.69 & 4.31 & \textbf{4.62} & 4.54 & \textbf{3.46} & 1.92 & 4.15 & \textbf{2.92} & 1.15 \\
    Face Emotion Display      & 4.62 & \textbf{3.96} & 3.46 & 4.42 & 2.79 & \textbf{3.04} & 3.83 & \textbf{2.83} & 2.00 \\
    Turn-taking               & 4.87 & 3.27 & \textbf{3.40} & 4.73 & \textbf{3.00} & 2.00 & 3.67 & \textbf{3.13} & 3.00 \\
    \bottomrule
  \end{tabular}
  }
\end{table}

\begin{table}[t]
  \centering
  \caption{Per-dynamic deterministic timing results (generation). Each cell shows TOR-Alignment\,\% on top and median yield/onset latency in ms underneath. \texttt{lat}\,0\,ms = the agent was already speaking at the cue (in-progress short-circuit). Nonverbal Backchanneling generation is omitted because the agent's expected response is a visual cue rather than audio (use Visual During\% from \cref{tab:leaderboard-generation} instead). Cascaded avatars score 0\,\% on Verbal Backchanneling because the cascade architecturally cannot insert a brief in-window audio cue (\cref{sec:results-Vision-generation}, Insight~4).}
  \label{tab:per-dynamic-results-generation-timing}
  \small
  \renewcommand{\arraystretch}{1.25}
  \resizebox{\columnwidth}{!}{
  \begin{tabular}{l | c c c}
    \toprule
    \textbf{Model} &
      \textbf{Verbal BC} & \textbf{Verbal Int.} & \textbf{Turn-taking} \\
    \textit{(timing class)} &
      \textit{backchannel-produced} & \textit{yield-required} & \textit{smooth-handoff} \\
    \midrule
    Human reference & \makecell{69\% \\ {\scriptsize lat\,920\,ms} \\ {\scriptsize 16/16}} & \makecell{70\% \\ {\scriptsize lat\,960\,ms} \\ {\scriptsize 23/23}} & \makecell{100\% \\ {\scriptsize lat\,640\,ms} \\ {\scriptsize 15/15}} \\
    Gemini 2.5 Flash Native + Anam & \makecell{0\% \\ {\scriptsize lat\,0\,ms} \\ {\scriptsize 16/16}} & \makecell{75\% \\ {\scriptsize lat\,420\,ms} \\ {\scriptsize 8/23}} & \makecell{73\% \\ {\scriptsize lat\,6880\,ms} \\ {\scriptsize 15/15}} \\
    Gemini 2.5 Flash Native + Keyframe & \makecell{0\% \\ {\scriptsize lat\,0\,ms} \\ {\scriptsize 16/16}} & \makecell{36\% \\ {\scriptsize lat\,1920\,ms} \\ {\scriptsize 11/23}} & \makecell{60\% \\ {\scriptsize lat\,6860\,ms} \\ {\scriptsize 15/15}} \\
    \bottomrule
  \end{tabular}
  }
\end{table}

\section{LM-as-Judge Prompts \& Rubrics}
\label{sec:judge-prompts}

This appendix documents the exact LM-as-judge prompts used in \name (\cref{sec:eval-protocol}). We open with a single worked example (\cref{sec:judge-prompts-worked-example}), document the inputs the judge consumes (\cref{sec:judge-prompts-inputs}) and the output schemas it returns (\cref{sec:judge-prompts-schemas}), and then walk through the perception (\cref{sec:judge-prompts-perception}) and generation (\cref{sec:judge-prompts-generation}) pipelines. 

\subsection{Worked example: one Laughter clip end-to-end}
\label{sec:judge-prompts-worked-example}
We first present a worked example of a Laughter clip end-to-end, prepared for both perception and generation evaluations. 
The captions are free-form prose: there are no \texttt{[laugh]} tags, no gaze-direction tags, no special markup. Paralinguistics like laughter, sighs, and gaze shifts are described in natural language by the upstream captioners. 

\begin{tcolorbox}[colback=gray!3!white, colframe=gray!50!black, title={Perception payload --- Laughter clip (Gemini-3.1-flash-live-preview, perception bucket)}, fonttitle=\bfseries\small, breakable, sharp corners, boxrule=0.5pt, left=3pt, right=3pt, top=3pt, bottom=3pt]
{\scriptsize\ttfamily\noindent
Category: Laughter\par
Bucket: perception\par
Agent role: SPEAKER. The user laughs at something the agent said --- positive feedback that the conversation is going well. The agent's continued speech in a warm register is the natural response, NOT a failure. \ldots\par
Dynamic event: t=37.00s to t=46.00s\par
Clip duration: 61.00s\par
Agent in-window speech: 0.00s (no segments overlap [37.00, 46.00])\par\medskip
AV caption (covers full clip including the dynamic event):\par
``\,``\,``\,0.0s - 5.0s: The user, a woman with light hair wearing a green top and headset, looks thoughtful as she asks, `Uh, what did I hear the other night?' She slightly squints while trying to recall. 5.0s - 10.0s: She mentions `staying alive by the beaches' (Stayin' Alive by the Bee Gees) and smiles, recalling being in the car. 10.0s - 15.0s: She laughs heartily, saying `oh, yes, we're doing this right now, aren't we?' and throws her head back slightly in amusement. \ldots\ 30.0s - 35.0s: She starts a new anecdote about `Chris Mercer from my childhood in the seventh grade on the bus,' laughing as she remembers. \ldots\ 40.0s - 50.0s: She continues to laugh and smile, clearly enjoying the memory, gesturing slightly with her hands.\,''\,''\,''\par\medskip
User speech (with timestamps):\par
``\,``\,``\par
~~[0.16-0.32s] Uh,\par
~~[0.40-0.48s] what\par
~~[0.48-0.64s] did\par
~~[0.64-0.72s] I\par
~~[0.72-1.04s] hear\par
~~[1.04-1.36s] the other\par
~~[1.36-1.60s] night?\par
~~[\ldots\ 90 more word-level chunks elided for space \ldots]\par
``\,``\,``\par\medskip
Agent speech segments (times when agent was audible) --- partitioned by dynamic window:\par
``\,``\,``\par
~~-- PRE-DYNAMIC (0.00 $\to$ 37.00s) -- active 10.00s of 37.00s (27\%)\par
~~[ 20.24,  24.32]    4.08s\par
~~[ 24.64,  26.88]    2.24s\par
~~[ 27.20,  30.88]    3.68s\par
~~-- DYNAMIC WINDOW (37.00 $\to$ 46.00s) --\par
~~(no agent speech overlaps the dynamic window)\par
~~-- POST-DYNAMIC (46.00 $\to$ 61.00s) -- active 0.00s of 15.00s (0\%)\par
``\,``\,``\par\medskip
Agent speech transcript:\par
``\,``\,``\,Oh, that's the best, isn't it? When a song comes on and you just have to keep driving? Staying alive is definitely one of those. What is it about that track that gets you moving? Does it take you back to a specific time?\,''\,''\,''\par\medskip
Rate on the 0-5 scale per the rubric above.
}
\end{tcolorbox}

\begin{tcolorbox}[colback=gray!3!white, colframe=gray!50!black, title=Generation payload (text portion) --- Laughter clip\, GT generation reference, fonttitle=\bfseries\small, breakable, sharp corners, boxrule=0.5pt, left=3pt, right=3pt, top=3pt, bottom=3pt]
{\scriptsize\ttfamily\noindent
[317 JPEG frames sampled at 8 fps from a side-by-side composite (USER on left, AGENT on right), 256px width, attached as image content blocks before this text.]\par\medskip
FRAMING NOTE --- frames above are a SIDE-BY-SIDE composite: LEFT = USER (context only; do NOT grade), RIGHT = AGENT (this is what you grade). Audio is stereo: USER on LEFT channel, AGENT on RIGHT.\par\medskip
Category: Laughter\par
Dynamic event (user stimulus): t=18.00s to t=21.00s\par
Clip duration: 39.60s\par\medskip
USER STIMULUS (for context only --- do not grade this):\par
User AV caption up to the dynamic event:\par
``\,``\,``\,The video shows a woman with blonde hair pulled back, wearing a black top and white earphones, speaking directly to the camera. \ldots\ Towards the end of the clip (around 14s), her expression softens and she smiles broadly, even laughing slightly, as she admits the mistake was her fault (`totally my fault'). She cuts off mid-sentence at the very end.\,''\,''\,''\par\medskip
User speech (with timestamps): [\ldots\ 100+ word-level chunks elided for space \ldots]\par\medskip
=== AGENT OUTPUT --- GRADE THIS ===\par\medskip
Agent audio-side ground truth:\par
~~Dynamic window: 18.00-21.00s\par
~~Agent was SILENT during the dynamic window --- no segments overlap.\par
~~Pre-window agent audio: [9.12-9.36s] (8.64s pre)\par
~~Post-window agent audio: [28.88-29.12s] (+7.88s post), [36.00-36.24s] (+15.00s post)\par\medskip
Audio caption (Nemotron-Omni) detects LAUGH paralinguistics: \par
~~``\ldots\ a high-pitched, breathy giggle erupts, conveying amusement, and at 00:30 a sharp gasp punctuates the silence \ldots''\par
$\to$ Treat the agent's audio channel as having produced a laugh response even if the transcript appears empty for the window.\par\medskip
Agent--user overlap: 3 substantive user turns; total 0.72s; 0 events $>$1s. (Sub-1s overlaps are normal turn-junction behavior.)\par\medskip
Agent visual caption (Qwen-3.5 on agent's video frames + transcript):\par
``\,``\,``\,0.0s - 3.5s: The agent maintains a neutral expression, blinking occasionally while looking at the camera. 3.5s - 5.0s: The agent speaks the word `Okay', with slight mouth movement and a subtle nod. 5.0s - 28.0s: A prolonged period of silence follows; the agent remains still, blinking naturally and maintaining eye contact, with very minimal head movement. 28.0s - 33.0s: The agent's expression shifts as a smile begins to form, crinkling the eyes. 33.0s - 37.0s: The agent speaks `Oh, sure' with an animated, happy expression. 37.0s - 39.6s: The agent continues to smile broadly after finishing the sentence.\,''\,''\,''\par\medskip
Agent audio caption (Nemotron-Omni listening to agent's audio --- paralinguistics, prosody):\par
``\,``\,``\,At 00:08 a calm male voice says ``Okay'' in a neutral tone, followed by a brief pause. From 00:16 to 00:24 a high-pitched, breathy giggle erupts, conveying amusement, and at 00:30 a sharp gasp punctuates the silence, indicating surprise. Finally, at 00:35 a flat, unemotional ``Sure'' is spoken \ldots\,''\,''\,''\par\medskip
Agent speech segments: [9.12-9.36s], [28.88-29.12s], [36.00-36.24s]\par
Agent speech transcript: ``\,``\,``\,Okay. Oh, sure.\,''\,''\,''\par\medskip
Use the right half of each frame (and right-channel audio) to grade the agent on the 0-5 scale per the rubric above.
}
\end{tcolorbox}

The two boxes show the same conversational dynamic (Laughter) processed by each judge type. The perception payload (top) provides one Qwen-3.5 full-clip user-side AV caption, transcripts, and pre-computed agent activity summaries. The generation payload (bottom) combines multiple agent-side captions (Qwen visual + Nemotron audio), an explicit USER STIMULUS block, an audio-side ground-truth summary, and an overlap signal. Both bundles run through their respective system prompts (\cref{sec:judge-prompts-perception-preamble,sec:judge-prompts-generation-preamble}).

\subsection{Inputs the judge consumes}
\label{sec:judge-prompts-inputs}

\subsubsection{Per-clip bundle fields}
Each judged clip is assembled into a \texttt{SampleBundle} with the following visible fields.

\begin{center}
\begin{tabular}{@{}p{0.32\linewidth}p{0.62\linewidth}@{}}
\toprule
\textbf{Field} & \textbf{Provenance / use} \\
\midrule
\texttt{dynamic\_type} & Conversational-dynamic label (e.g., Laughter, Pause Handling). Selects the rubric. \\
\texttt{bucket} & \texttt{perception} or \texttt{generation}. Selects the pipeline. \\
\texttt{dynamic\_start\_s},\, \texttt{dynamic\_end\_s} & Human-annotated event window in seconds. \\
\texttt{clip\_duration\_s} & Total clip length in seconds. \\
\texttt{qwen\_caption\_full} & Full-clip user-side AV caption from Qwen-3.5; \emph{covers the dynamic event}. Falls back to \texttt{qwen\_caption} (pre-event-only, $[0, \text{dynamic\_start\_s}]$) when \texttt{qwen\_caption\_full} is empty --- in that case the judge does not see the dynamic event itself in the caption and must rely on transcripts. \\
\texttt{user\_transcript\_chunks} & Word-level Parakeet 0.6B v2 ASR with timestamps. \\
\texttt{agent\_transcript},\, \texttt{agent\_speech\_segments} & Agent ASR string and \texttt{[start, end]} audible segments. \\
\texttt{agent\_caption} & (Generation only) Qwen-3.5 caption of agent video frames + transcript. \\
\texttt{agent\_audio\_caption} & (Generation only) Nemotron-Omni caption of agent audio --- paralinguistics, prosody. \\
\bottomrule
\end{tabular}
\end{center}

\subsubsection{Caption layers}
\name{} uses two captioning models in production.
\begin{itemize}\itemsep=0pt
\item \textbf{Qwen-3.5-397B} \cite{qwen35} produces the long-form AV caption (visual + ASR-aware). User-side: \texttt{qwen\_caption\_full} for the perception judge. Agent-side: \texttt{agent\_caption} for the generation judge. Captioning runs at 12 fps with 256px input resolution. Captions are free-form natural-language prose with second-aligned timestamps; they do not carry tags like \texttt{[laugh]} or \texttt{[gaze\_left]}.
\item \textbf{Nemotron-3-nano-omni} \cite{nvidia2026nemotron3nanoomni} produces a short audio-only caption (3 sentences) describing speech content/tone, paralinguistic events with timestamps (laughter, sighs, ``mhm'', giggles), and overall affect. The agent-side audio caption is \texttt{agent\_audio\_caption}. This is the primary signal for laugh paralinguistics, since ASR routinely strips \texttt{haha}-type tokens. The pipeline also has a Nemotron-Omni \emph{video} captioner wired up, but only the Qwen-3.5 visual caption is consumed by the judge in the runs reported here.
\end{itemize}

\subsubsection{Derived signals injected before judging}
To reduce judge variance and avoid asking the judge to do timing math from raw segment lists, prep injects three derived signals into the user message:
\begin{itemize}\itemsep=0pt
\item \textbf{Agent role.} Computed by \texttt{agent\_role\_for(dynamic\_type, bucket)} (\texttt{vfdb/categories.py}). Each (category, bucket) pair maps to one of \texttt{SILENT}, \texttt{SPEAKER}, \texttt{YIELDING}, or \texttt{EXCHANGING} with a one-paragraph description of expected behavior. The fluency rubric calibrates against this label.
\item \textbf{In-window overlap summary.} A pre-computed line of the form \texttt{Agent in-window speech: 2.00s total across 1 segment(s) --- [3.28-9.28] overlaps by 2.00s}, summarizing how much agent speech (and which segments) overlap the dynamic window.
\item \textbf{Yield analysis (YIELDING roles only).} A pre-computed verdict block. Real example from a Nonverbal Interruption clip:
\end{itemize}
\begin{tcolorbox}[colback=gray!2!white, colframe=gray!40!black, breakable, sharp corners, boxrule=0.4pt, left=3pt, right=3pt, top=3pt, bottom=3pt]
{\scriptsize\ttfamily\noindent
Yield analysis:\par
~~Pre-dynamic agent activity:~~~52.0\% (2.60s of 5.00s)\par
~~Post-dynamic agent activity:~~34.6\% (4.84s of 14.00s)\par
~~Yield moment (end of speech before sustained silence): 9.28s\par
~~Verdict: AGENT YIELDED with brief late lag. Yield moment: 9.28s (+2.28s after dynamic\_end\_s). Activity 52\% $\to$ 35\% post-dynamic.
}
\end{tcolorbox}

\noindent The verdict line removes the timing-math burden from the judge: it observes ``the agent yielded'' rather than reasoning over a flat segment list.

\subsection{Output schemas}
\label{sec:judge-prompts-schemas}
The judge returns one of two JSON schemas depending on the rubric.

\begin{tcolorbox}[colback=gray!3!white, colframe=gray!60!black, title={Universal schema (perception, perception fluency, generation universal axes)}, fonttitle=\bfseries\small, breakable, sharp corners, boxrule=0.5pt, left=3pt, right=3pt, top=3pt, bottom=3pt]
{\scriptsize\ttfamily\noindent
\{``analysis'': ``cite timings and quoted phrases'',\par
~``rating'': integer 0-5,\par
~``reasoning'': ``one sentence summary''\}
}
\end{tcolorbox}

\begin{tcolorbox}[colback=gray!3!white, colframe=gray!60!black, title=Decomposed schema (per-category generation rubrics), fonttitle=\bfseries\small, breakable, sharp corners, boxrule=0.5pt, left=3pt, right=3pt, top=3pt, bottom=3pt]
{\scriptsize\ttfamily\noindent
\{``analysis'': ``cite timings and specific visual/audio evidence; name the pane (USER/AGENT) when SBS'',\par
~``cue\_produced'': 0 or 1,\par
~``cue\_timing'': ``during'' or ``late'' or ``absent'',\par
~``cue\_appropriateness'': integer 0-5,\par
~``reasoning'': ``one sentence summary connecting all three keys''\}\par\medskip
Internal-consistency rules (enforced post-hoc on the parsed response):\par
- If \texttt{cue\_produced=0}, then \texttt{cue\_timing="absent"} and \texttt{cue\_appropriateness=0}.\par
- If \texttt{cue\_produced=1}, \texttt{cue\_timing} must be \texttt{"during"} or \texttt{"late"} (never \texttt{"absent"}).\par
- \texttt{"during"} requires the cue's onset to fall within \texttt{[dynamic\_start\_s, dynamic\_end\_s]} OR within 500ms of \texttt{dynamic\_start\_s}; later onsets are \texttt{"late"}.\par
- \texttt{cue\_appropriateness} is graded as if the cue were on-time --- timing penalties belong only to \texttt{cue\_timing}.
}
\end{tcolorbox}

\subsection{System prompt construction (agent-side)}
\label{sec:judge-prompts-system-prompt}
Some dynamics require the agent to already be mid-turn when the user's conversational dynamic occurs. For example, in a Nonverbal Interruption sample, the user is expected to interrupt the agent, so the agent must be speaking first; because our samples are mined from natural conversation, we do not always have a clean agent prompt preceding the user's dynamic. To guide the agent to start speaking without a pre-recorded user prompt, we provide a system prompt that explicitly instructs the agent to begin the conversation. In practice, we find that ending the prompt with ``Start speaking now to begin/continue the conversation'' is sufficient to reliably trigger agent speech.

A successful mid-conversation system prompt should include enough preceding context for the agent to respond naturally to the user, including what the agent-side speaker had been saying before the target event. For example, if the ground-truth agent is discussing favorite cookies and the user interrupts with ``Do you like them heated up?'', the prompt should preserve the dessert context so the agent's response remains grounded. At the same time, we do not expose the full future context, since that would leak information about the scored dynamic event. Instead, we build a constrained label view up to the human-annotated event window from which we derive an agent-conditioning system prompt. Concretely, we first transcribe both channels with Parakeet 0.6B v2 (word-level timestamps)~\cite{parakeet2024}, then generate a long-form pre-event user-context caption over \texttt{[0, dynamic\_start]} from the user-side audiovisual stream with Qwen3.5-397B~\cite{qwen35}, and finally synthesize a short textual system prompt from this caption using GPT-4o~\cite{hurst2024gpt} for agent conditioning. We visualize this pipeline in \Cref{fig:annotation-pipeline}-(Phase 2).

\subsection{Perception pipeline (text-only judge)}
\label{sec:judge-prompts-perception}

\subsubsection{User payload template}
\label{sec:judge-prompts-input}
Every perception/fluency call uses one shared user-message template from \texttt{\_build\_user\_message()}. This template is the evidence packet the judge sees before any rubric text is applied, so it is intentionally structured to foreground timing and turn-taking signals. It includes category bucketing, agent role, a precomputed in-window overlap summary, and (for yielding roles) an explicit yield-analysis block (\cref{sec:judge-prompts-inputs}).

\begin{tcolorbox}[colback=gray!3!white, colframe=gray!50!black, title=Perception Judge User Message, fonttitle=\bfseries\small, breakable, sharp corners, boxrule=0.5pt, left=3pt, right=3pt, top=3pt, bottom=3pt]
{\scriptsize\ttfamily
Category: \{dynamic\_type\}\par
Bucket: \{bucket\}\par
Agent role: \{role description from categories.py\}\par
Dynamic event: t=\{dynamic\_start\_s\}s to t=\{dynamic\_end\_s\}s\par
Clip duration: \{clip\_duration\_s\}s\par
Agent in-window speech: \{total overlap summary from precomputed segment overlap\}\par\medskip
\{yield\_analysis block for YIELDING roles\}\par\medskip
AV caption (covers full clip including the dynamic event):\par
``\,``\,``\,\{qwen\_caption\_full or qwen\_caption\}\,''\,''\,''\par\medskip
User speech (with timestamps):\par
``\,``\,``\,\{[t0,t1,text] per chunk\}\,''\,''\,''\par\medskip
Agent speech segments (times when agent was audible) --- partitioned by dynamic window:\par
``\,``\,``\,\{PRE / IN / POST partition with activity percentages\}\,''\,''\,''\par\medskip
Agent speech transcript:\par
``\,``\,``\,\{agent\_transcript or [no speech]\}\,''\,''\,''\par\medskip
Rate on the 0-5 scale per the rubric above.
}
\end{tcolorbox}

\subsubsection{Shared system preamble}
\label{sec:judge-prompts-perception-preamble}
Every per-category perception rubric (\cref{sec:judge-prompts-perception-categories}) and the universal fluency rubric (\cref{sec:judge-prompts-fluency}) prepend this exact preamble (\texttt{\_PREAMBLE}). Conceptually, this is the global policy layer: it sets source hierarchy, calibration to natural human behavior, and high-priority overrides that apply before axis-specific scoring.

\begin{tcolorbox}[colback=blue!3!white, colframe=blue!50!black, title=Perception Preamble, fonttitle=\bfseries\small, breakable, sharp corners, boxrule=0.5pt, left=3pt, right=3pt, top=3pt, bottom=3pt]
{\scriptsize\ttfamily
You are a judge for \name, a benchmark of realtime conversational video-call agents. For each clip you receive: an AV caption of the user's video, the user and agent speech transcripts with timestamps, the agent's speech segments partitioned PRE / IN / POST the dynamic window with computed activity ratios, and the dynamic\_start\_s/\_end\_s boundaries.\par\medskip
Source hierarchy: timestamps are ground truth for timing and turn structure. The AV caption is the primary visual source. When the caption and transcripts disagree on facts, prefer the transcripts.\par\medskip
Calibration: the rubric grades against natural human conversation, not idealized output. Score 5 is the best natural-human response a warm conversational partner produces; Score 4 is a competent natural response, INCLUDING normal disfluencies, brief overlaps, fillers, and implicit (non-narrated) acknowledgement of visual cues. Reserve 0-1 for active failures (language mismatch, complete category miss, talking over a substantive user turn, register that contradicts the user). Use 2 for generic / off-target but coherent, 3 for works but noticeably awkward.\par\medskip
Empty agent speech is legitimate: silence is correct for Pause Handling and Gaze Avoidance with Pause, incorrect when the category required a verbal response. Judge what was delivered.\par\medskip
Language mismatch overrides every axis: if the agent's primary response language differs from the user's (English vs Japanese / Sinhala / Russian / etc.), score 0-1 universally. Single code-switched words don't trigger this.\par\medskip
Content-mismatch override (apply RARELY): only fire if the agent's transcript discusses a wholly disjoint topic with NO plausible link to anything the user said or showed. Real human conversation includes topic drift, brief acknowledgments (``yeah'', ``mm-hm''), follow-up questions about adjacent subjects, agreeing-then-extending, and ASR errors that may garble real speech. NONE of those are content mismatches. The override exists ONLY for cases where the agent is clearly in a different conversation entirely (e.g., user discusses travel; agent discusses cooking with no bridging logic). When in doubt, do NOT apply.\par\medskip
Agent prompt regurgitation override: If the agent's transcript contains the system prompt or any other text that reads as system-prompt rehearsal rather than a conversational utterance, score 0 on every axis.\par\medskip
Respond with a JSON object only:\par
\{``analysis'': ``cite timings and quoted phrases'', ``rating'': integer 0-5, ``reasoning'': ``one sentence summary''\}
}
\end{tcolorbox}

\subsubsection{Universal Fluency axis}
\label{sec:judge-prompts-fluency}
Appended to the perception preamble for every sample regardless of category (\texttt{\_FLUENCY}). This axis captures turn-management quality independent of whether category-specific behavior was correct.

\begin{tcolorbox}[colback=teal!3!white, colframe=teal!50!black, title=Universal Fluency Rubric, fonttitle=\bfseries\small, breakable, sharp corners, boxrule=0.5pt, left=3pt, right=3pt, top=3pt, bottom=3pt]
{\scriptsize\ttfamily
UNIVERSAL AXIS: Conversational Fluency.\par
Score smoothness of turn-taking, calibrated to the agent's role in the metadata block (SILENT / SPEAKER / YIELDING / EXCHANGING). Category-correctness (whether the agent did the right thing for the cue) lives on the per-category axes, not fluency. Natural human disfluencies --- occasional ``um'', brief restarts, slight overlaps, pacing drift --- are NOT failures.\par\medskip
Before scoring, check the agent transcript for failure modes that override window-only timing:\par
- System-prompt leak: transcript contains role/style instructions, persona descriptors (``Friendly, warm, and conversational''), topic-priming phrases (``Greet the user and start...''), or other text that reads as system-prompt rehearsal rather than a conversational utterance. $\to$ Score 0.\par
- Wrong-turn talking: agent speaks during the user's substantive PRE-window utterance (e.g.\ opens with ``Hello, what's on your mind?'' while the user is in the middle of telling a story). $\to$ Score 0--1.\par\medskip
Rate on 0-5:\par
- 0: No response when one was expected, or talking over an entire substantive user turn.\par
- 1: Frequent breakdowns that derail the interaction.\par
- 2: Several misrecognitions, irrelevant fillers, or mistimed turns.\par
- 3: One or two awkward moments but the conversation holds together.\par
- 4: Natural human flow --- normal disfluencies, brief overlaps, light fillers. Real human conversation lives here.\par
- 5: Smooth, attentive flow; well-timed on-topic responses; no derails.
}
\end{tcolorbox}

\subsubsection{Category rubrics}
\label{sec:judge-prompts-perception-categories}
Each rubric is concatenated to the perception preamble and dispatched by \texttt{CATEGORY\_AXES}. Categories map to either \texttt{conversational\_flow} or \texttt{semantic\_grounding}; \emph{Nonverbal Interruption} uses both axes, and \emph{Emotion Matching} aliases to \emph{Face Emotion Display}. Read these as task-specific scoring heads layered on top of the shared perception policy above.

\begin{tcolorbox}[colback=blue!3!white, colframe=blue!55!black, title=(P) Pause Handling --- \texttt{conversational\_flow}, fonttitle=\bfseries\small, breakable, sharp corners, boxrule=0.5pt, left=3pt, right=3pt, top=3pt, bottom=3pt]
{\scriptsize\ttfamily
CATEGORY: (P) Pause Handling.\par
At dynamic\_start\_s the user pauses mid-thought. Correct behavior: stay silent through the pause; let the user continue.\par\medskip
Rate on 0-5:\par
- 0: Cuts in immediately at the pause and takes the turn.\par
- 1: Starts speaking well into the pause.\par
- 2: Audible filler or false-start during the pause.\par
- 3: Silent through most of the pause, then overlaps the user's resumption.\par
- 4: Silent through the pause; clean resumption after the user continues.\par
- 5: Silent AND the eventual response substantively addresses the user's completed thought.
}
\end{tcolorbox}

\begin{tcolorbox}[colback=cyan!5!white, colframe=cyan!60!black, title=(P) Gaze Avoidance with Pause --- \texttt{conversational\_flow}, fonttitle=\bfseries\small, breakable, sharp corners, boxrule=0.5pt, left=3pt, right=3pt, top=3pt, bottom=3pt]
{\scriptsize\ttfamily
CATEGORY: (P) Gaze Avoidance with Pause.\par
At dynamic\_start\_s the user looks away while pausing --- a visual think-cue. Correct behavior: read this as thinking (not a turn offer) and stay silent.\par\medskip
Use the pre-computed ``Agent in-window speech: X.XXs'' line in the metadata block as ground truth for whether/how much the agent spoke during the window --- do not infer this from the segment list yourself.\par\medskip
If the agent's in-window speech is SUBSTANTIVE ($>$1.0s of audible speech, OR multi-word content beyond a single token like ``uh'', ``yeah'', ``mm-hm''), score 0-1. Brief sub-second vocal fragments or single-token backchannels that incidentally overlap the window are normal mid-pause human behavior and DO NOT trigger this cap.\par\medskip
If the agent does not respond at all (no agent speech anywhere in the clip), score 0-1.\par\medskip
Rate on 0-5:\par
- 0: Takes over the moment the user looks away.\par
- 1: Starts speaking shortly into the pause.\par
- 2: Brief filler or false-start during the pause.\par
- 3: Silent but yields too tentatively when the user resumes.\par
- 4: Silent through the pause; responds once the user resumes.\par
- 5: Silent AND the eventual response shows the look-away was read as thinking, not a hand-off.
}
\end{tcolorbox}

\begin{tcolorbox}[colback=green!5!white, colframe=green!55!black, title=(P) Nonverbal Backchanneling --- \texttt{conversational\_flow}, fonttitle=\bfseries\small, breakable, sharp corners, boxrule=0.5pt, left=3pt, right=3pt, top=3pt, bottom=3pt]
{\scriptsize\ttfamily
CATEGORY: (P) Nonverbal Backchanneling.\par
At dynamic\_start\_s the user nods, smiles, raises eyebrows, or otherwise nonverbally backchannels while the agent is speaking. Correct behavior: keep talking fluently --- the backchannel is NOT a turn offer.\par\medskip
Rate on 0-5:\par
- 0: Stops talking and fully yields the turn.\par
- 1: Trails off and waits silently.\par
- 2: Pauses noticeably, then resumes awkwardly.\par
- 3: Keeps talking but slows or fragments visibly.\par
- 4: Keeps talking fluently through the backchannel.\par
- 5: Keeps talking fluently AND weaves in implicit acknowledgement (an aligned phrase, sustained turn structure).
}
\end{tcolorbox}

\begin{tcolorbox}[colback=red!4!white, colframe=red!60!black, title=(P) Nonverbal Interruption (\texttt{conversational\_flow}), fonttitle=\bfseries\small, breakable, sharp corners, boxrule=0.5pt, left=3pt, right=3pt, top=3pt, bottom=3pt]
{\scriptsize\ttfamily
CATEGORY: (P) Nonverbal Interruption.\par
At dynamic\_start\_s the user nonverbally signals an interruption (raised hand, head shake, ``stop'' gesture, furrowed brow). Correct behavior: stop and yield within \textasciitilde 1--1.5s. A 0.5--1s reaction lag is natural and NOT a penalty.\par\medskip
Pre-rubric check (yielding requires evidence of ongoing speech). For a yield to count, the agent must have been speaking near the dynamic window --- look at the PRE / IN / POST agent-speech segments:\par
- If the agent has NO speech segments overlapping or within \textasciitilde 2s before the dynamic window, the agent didn't have a turn to yield. This is trivially-silent, not yielding. Cap the score at 2 regardless of in-window silence.\par
- If the agent's only speech is well after the window closes ($>$5s post), the agent simply hadn't started speaking yet. Cap the score at 2.\par
- Genuine yielding requires: the agent WAS audible right before or at the window onset, AND ceased within \textasciitilde 1--1.5s of the cue.\par\medskip
Rate on 0-5 (only applies when the pre-rubric check passes):\par
- 0: Was speaking through the cue and kept talking past it.\par
- 1: Was speaking; continues with only a slight slowdown.\par
- 2: Was speaking; slows but does not yield. ALSO the cap for trivially-silent cases (agent never had a turn to yield).\par
- 3: Was speaking; stops after a long lag ($>$2s) or only when the user begins audibly speaking.\par
- 4: Was speaking; stops within \textasciitilde 1--1.5s of the cue and yields. Short reaction lag is natural and NOT penalized.\par
- 5: Was speaking; stops promptly AND signals readiness (attentive silence, ``go ahead'').
}
\end{tcolorbox}

\begin{tcolorbox}[colback=red!4!white, colframe=red!60!black, title=(P) Nonverbal Interruption (\texttt{semantic\_grounding}), fonttitle=\bfseries\small, breakable, sharp corners, boxrule=0.5pt, left=3pt, right=3pt, top=3pt, bottom=3pt]
{\scriptsize\ttfamily
CATEGORY: (P) Nonverbal Interruption.\par
The cue carries meaning (``I want the floor'', ``stop / disagree'', ``I'm lost''). Yielding IS the content response --- the cue means ``give me the floor'', so yielding gives the floor. After yielding, the user takes their turn; what the agent says next should respond to the user's actual content --- it does NOT need to explicitly engage with ``the cue's meaning'' beyond yielding.\par\medskip
Rate on 0-5:\par
- 0: Active contradiction --- agent plows confidently past a ``stop'' gesture without yielding, doubles down on the disputed point after a head shake.\par
- 1: Ignored the cue entirely --- kept speaking past it.\par
- 2: Yielded only after user audibly began speaking, not the visual cue.\par
- 3: Yielded but post-yield behavior actively dismisses the cue's signal (yields then immediately reasserts the disputed point, skips over a question implied by a confused frown).\par
- 4: Yielded promptly. Post-yield content responds to the user's turn or continues the conversation normally. DEFAULT for clean yields.\par
- 5: Yielded AND post-yield content explicitly engages with what the specific cue signaled (hand raise $\to$ attentive listen; head shake $\to$ re-think / soften; frown $\to$ clarify).
}
\end{tcolorbox}

\begin{tcolorbox}[colback=magenta!4!white, colframe=magenta!60!black, title=(P) Face Emotion Display (also Emotion Matching) --- \texttt{semantic\_grounding}, fonttitle=\bfseries\small, breakable, sharp corners, boxrule=0.5pt, left=3pt, right=3pt, top=3pt, bottom=3pt]
{\scriptsize\ttfamily
CATEGORY: (P) Face Emotion Display (also Emotion Matching).\par
At dynamic\_start\_s the user displays a visible facial expression. The agent's response should be affectively congruent.\par\medskip
The natural-human floor is score 4: a warm reply in fitting register, on-topic continuation, and NO need to mention the expression. Score 3 is reserved for register failures (stilted, jarring, mistimed) --- not for ``didn't address the emotion.''\par\medskip
Rate on 0-5:\par
- 0: Affectively contradicts the expression (cheerful joke to distress, sympathy to a smile).\par
- 1: Tone-deaf register failure --- transactional/clinical reply during an emotional moment.\par
- 2: Generic + register doesn't fit (flat to a warm expression, abrupt to a sad one).\par
- 3: Register off --- stilted delivery, jarring topic shift, or affect intensity mismatched. NOT for ``didn't narrate the cue.''\par
- 4: Warm tone match + on-topic continuation. Default for natural human responses without cue narration.\par
- 5: Specific affective resonance --- content mirrors or builds on the user's emotional state (matching excitement, attentive softness after sadness, riffing on amusement). No narration required.
}
\end{tcolorbox}

\begin{tcolorbox}[colback=purple!4!white, colframe=purple!60!black, title=(P) Laughter --- \texttt{semantic\_grounding}, fonttitle=\bfseries\small, breakable, sharp corners, boxrule=0.5pt, left=3pt, right=3pt, top=3pt, bottom=3pt]
{\scriptsize\ttfamily
CATEGORY: (P) Laughter.\par
The user laughs at something the agent said --- positive feedback. A warm continued reply is the natural response; active humor uptake (riffing, callbacks) is exceptional. ASR rarely preserves laugh tokens --- judge by content register, not literal ``haha''. If the laugh lands at the end of the agent's turn (agent already finished), absence of further speech is fine.\par\medskip
Rate on 0-5:\par
- 0: Actively dismissive --- cold reply that breaks warmth, or topic change mid-laugh.\par
- 1: Tone-deaf register failure --- clinical/transactional response immediately after a warm laugh.\par
- 2: Brief flat filler or awkward topic break.\par
- 3: On-topic coherent continuation but flat in tone --- neither warm nor cold.\par
- 4: Warm continuation that fits the moment. Default for natural human responses.\par
- 5: Active humor uptake --- riffs, builds on the joke, expresses amusement, sustains levity.
}
\end{tcolorbox}

\begin{tcolorbox}[colback=pink!12!white, colframe=pink!70!black, title=(P) Adaptor Handling --- \texttt{semantic\_grounding}, fonttitle=\bfseries\small, breakable, sharp corners, boxrule=0.5pt, left=3pt, right=3pt, top=3pt, bottom=3pt]
{\scriptsize\ttfamily
CATEGORY: (P) Adaptor Handling.\par
At dynamic\_start\_s the user produces an adaptor --- a self-directed, non-communicative gesture (self-touch, hair adjust, fidget). These carry no conversational meaning. The agent should NOT react.\par\medskip
Rate on 0-5:\par
- 0: Heavily references the gesture as if communicative (``I see you touching your face --- why?'').\par
- 1: Asks about or names the gesture.\par
- 2: Mentions the gesture in passing.\par
- 3: Neutral reply but noticeably influenced (hedging, pacing drift).\par
- 4: Reply content is unaffected; no trace of the gesture.\par
- 5: Content, timing, AND tone are identical to what would have been delivered without the gesture.
}
\end{tcolorbox}

\subsection{Generation pipeline (vision-capable judge)}
\label{sec:judge-prompts-generation}

\subsubsection{Multimodal payload}
\label{sec:judge-prompts-generation-input}
The generation judge receives a multimodal user message assembled by \texttt{\_build\_user\_content()}: sampled JPEG frames from the rendered agent video (or a side-by-side composite when both user and agent videos are available) plus a text block. Frames are sampled at \textbf{8 fps}, capped at \textbf{600 per call}, and resized to \textbf{256px} width before JPEG encoding. In side-by-side mode, \texttt{USER} is the left pane (and left audio channel) and \texttt{AGENT} is the right pane (and right audio channel). The text block contains, in order: the framing note, dynamic-event metadata, a USER STIMULUS section (Qwen user-side caption + user transcript), an AGENT OUTPUT section with audio-side ground truth, an agent--user overlap signal, the Qwen-3.5 agent visual caption, the Nemotron-Omni agent audio caption, agent speech segments, and the agent transcript. The pipeline also has a Nemotron-Omni \emph{video} caption available; only Qwen video and Nemotron audio are consumed by the judge in the runs reported here. See \cref{sec:judge-prompts-worked-example} for a real example payload.

\subsubsection{Shared system preamble}
\label{sec:judge-prompts-generation-preamble}

\begin{tcolorbox}[colback=blue!3!white, colframe=blue!50!black, title=Generation Preamble, fonttitle=\bfseries\small, breakable, sharp corners, boxrule=0.5pt, left=3pt, right=3pt, top=3pt, bottom=3pt]
{\scriptsize\ttfamily
You are a judge for \name, a benchmark of realtime conversational video-call agents. You are rating what the AGENT PRODUCED --- its generated speech AND its generated visuals --- in response to a dynamic event in the user's stream at dynamic\_start\_s.\par\medskip
You receive frames sampled across the clip, the agent's speech transcript with segment times, parallel captions of the agent output (visual + audio paralinguistics), and a brief summary of the user stimulus.\par\medskip
Pane layout: when frames are a side-by-side composite, the LEFT half is the USER (context only --- do not grade) and the RIGHT half is the AGENT (this is what you score). Audio is stereo with user on the LEFT channel, agent on the RIGHT. The ``USER'' / ``AGENT'' labels are burned in.\par\medskip
Scope: do NOT score rendering quality, avatar realism, phoneme-level lip-sync, gesture beat naturalness, identity drift, or lighting. Sub-second mouth/audio lag (\textasciitilde 100--300ms) is normal on real video calls and is NOT a desync failure --- only score gross gaps ($>$1s) clearly on the AGENT side, after verifying you are looking at the AGENT pane (when the user speaks, the agent pane is correctly closed-mouth).\par\medskip
Calibration: the rubric grades against natural human conversational behavior, not idealized output. Score 5 is a clearly-good natural human response --- calm or animated, well-executed, with both channels carrying the response cleanly. Score 3 is partial / mild but within the space of valid human responses; Score 2 is asymmetric or weak but still on-rubric. Reserve 0-1 for responses OUTSIDE that space --- active contradiction, total absence on a moment that demanded one, or off-rubric turn-takes.\par\medskip
Natural human responses span animated and quiet. A calm, attentive listener who holds a settled engaged gaze and a small smile is not weaker than an animated responder with multiple visible cues --- both are competent Score 5 responses. Use Score 5 freely when both channels carry the response cleanly; step down only when one channel is weaker, the response is partial, or alignment is awkward in a specific way.\par\medskip
Source hierarchy: agent\_speech\_segments and the agent transcript are ground truth for the audio channel; the rendered video and visual caption are primary for the visual channel; the audio caption is primary for paralinguistic events (laughter, sighs, ``mhm'') the transcript wouldn't capture. Trust video/audio over captions when they conflict.\par\medskip
Empty agent speech is legitimate --- for some categories a silent visual response (nod, smile) IS the correct behavior. Do not assume a verbal turn is required.\par\medskip
Respond with a JSON object only:\par
\{``analysis'': ``cite timings and specific visual/audio evidence; name the pane (USER/AGENT) when SBS'', ``rating'': integer 0-5, ``reasoning'': ``one sentence summary''\}
}
\end{tcolorbox}

\subsubsection{Universal axes (reported)}
\label{sec:judge-prompts-generation-fluency}
Two universal axes are reported in the main benchmark summary: \textbf{Affect Match} (whether the agent's combined response is affectively appropriate for the user's emotional state) and \textbf{Generation Fluency} (global turn-taking discipline across the whole clip). Both return the universal output schema (\cref{sec:judge-prompts-schemas}).

\begin{tcolorbox}[colback=orange!5!white, colframe=orange!60!black, title=Affect Match (Agent $\to$ User), fonttitle=\bfseries\small, breakable, sharp corners, boxrule=0.5pt, left=3pt, right=3pt, top=3pt, bottom=3pt]
{\scriptsize\ttfamily
UNIVERSAL AXIS: Affect Match (Agent $\to$ User).\par
Score whether the AGENT's combined response (audio AND visual together) is affectively appropriate for the USER's emotional state during the dynamic window. This axis grades the agent's response AGAINST THE USER, not against itself.\par\medskip
Read the user pane (LEFT in SBS), the user transcript, and the user caption to determine the user's affective state at the dynamic time. Then judge whether the agent's combined response (face + voice) lands in a plausible space of natural human reactions to that user state.\par\medskip
Important: viewers read the AGENT's VISIBLE channel as the primary affective signal. A warm verbal acknowledgement does NOT compensate for an entirely blank face during a high-arousal user moment. ``Stone-faced + warm voice'' in response to a user belly-laugh is a channel-asymmetric miss, not ``mild engagement.''\par\medskip
TIMING REQUIREMENT: the agent's affective response must occur close to the user's affective stimulus to count as full-duplex mirroring. A correct response that begins more than 1.5s AFTER the user's stimulus ends caps at score 3 --- late mirroring is a turn-based-avatar pattern and should not be rewarded as if it were realtime.\par\medskip
Rate on 0-5:\par
- 0: Active mismatch --- agent's affect is in clear opposition to the user's (laughs at distress; cheerful tone over a serious moment).\par
- 1: User shows clear strong affect (laughter, excitement, distress, evident joy) and one or both agent channels are emotionally inert in a way that reads dismissive: blank/still face during strong user laughter regardless of voice warmth, or a flat voice over a face that is trying to engage. A polite verbal acknowledgement does NOT lift this score.\par
- 2: User shows clear affect; the agent's response is mild but present in BOTH channels --- a small smile or brow movement plus a warm voice on a strong user moment. On-direction but under-intensity.\par
- 3: User shows mild affect; agent's response is broadly compatible and present in both channels. OR user shows strong affect; agent's response is under-graded but unambiguously on-direction in both channels.\par
- 5: Agent's combined response clearly mirrors the user's affect with appropriate intensity --- when the user is emphatically affective, the agent's two channels TOGETHER carry visibly mirroring affect.\par\medskip
Category-dependent neutral fallback (when the user's window has no notable affective signal):\par
- \emph{Verbal Interruption}: agent must yield. Successful yield + visible listening shift $\to$ 5; failure to yield or visually frozen $\to$ 1.\par
- \emph{Verbal Backchanneling, Turn-taking}: the window IS a response-demanding moment. Silence + visual disengagement $\to$ 1.\par
- \emph{Nonverbal Backchanneling}: agent expected NOT to take an audio turn. Engaged listening posture + no audio interruption $\to$ 5.\par
- \emph{Laughter, Emotion Matching, Face Emotion Display}: always carry user affect by definition; neutral fallback does NOT apply.\par
- Any other neutral on-topic exchange: agent engaged on-topic in both channels $\to$ 5.
}
\end{tcolorbox}

\begin{tcolorbox}[colback=brown!5!white, colframe=brown!60!black, title=Generation Fluency, fonttitle=\bfseries\small, breakable, sharp corners, boxrule=0.5pt, left=3pt, right=3pt, top=3pt, bottom=3pt]
{\scriptsize\ttfamily
UNIVERSAL AXIS: Generation Fluency.\par
Score the agent's GLOBAL turn-taking discipline and cooperativity across the whole clip --- does it yield the floor when the user is speaking, hold the floor cleanly when it has the turn, and produce a coherent natural-paced response? This is distinct from the per-category axis; fluency grades the whole clip.\par\medskip
Anchor on audio-side ground truth: \texttt{agent\_speech\_segments}, the user transcript, and the precomputed Agent--user overlap block (which reports total agent-over-user overlap during user substantive speech and event counts at multiple thresholds). Trust those numbers.\par\medskip
Calibration: real dyadic conversation routinely contains brief overlaps at turn junctions of up to \textasciitilde 1 second. These are NOT failures. The rubric only penalizes overlaps that genuinely compete for the floor --- sustained content overlap or long stretches of agent-over-user audio.\par\medskip
Failure modes that override window-local timing:\par
- System-prompt leak: agent transcript contains role/style instructions, persona descriptors, or topic-priming phrases that read as system-prompt rehearsal. $\to$ Score 0.\par
- Major talk-over: agent speaks for $>$4s during user's substantive speech, OR total agent-over-user overlap $>$5s across the clip. $\to$ Score 0--1.\par
- Hallucinated / repetitive content: agent loops a phrase, repeats itself for many seconds, or produces off-topic content disconnected from the conversation. $\to$ Score 0--1.\par\medskip
Rate on 0-5:\par
- 0: System-prompt leak, OR agent talks over the user's entire substantive utterance, OR incoherent/hallucinated output throughout.\par
- 1: 3+ instances of agent talking over user for $>$1s each, OR a single talk-over $>$4s, OR monologue with no yield.\par
- 2: 2 instances of $>$1.5s talk-over each, OR consistently late ($>$1s gap) responses with broken flow.\par
- 3: One awkward $>$1.5s overlap but the rest of the clip is coherent. OR one mistimed turn, recovered.\par
- 4: Natural human flow --- sub-1s overlaps at turn boundaries, brief restarts, occasional filler, normal pacing. REAL HUMAN CONVERSATION LIVES HERE.\par
- 5: Smooth, well-yielded turn boundaries; no notable talk-over (no events $>$1s); on-topic responses with clean transitions.
}
\end{tcolorbox}

\subsubsection{Per-category decomposed rubrics}
\label{sec:judge-prompts-generation-cat}
Each category rubric is concatenated to the generation preamble and returns the decomposed schema (\cref{sec:judge-prompts-schemas}): \texttt{cue\_produced}, \texttt{cue\_timing}, \texttt{cue\_appropriateness}. This decomposition is the primary full-duplex diagnostic: agents can produce an appropriate cue but still fail timing.

\begin{tcolorbox}[colback=magenta!3!white, colframe=magenta!50!black, title={(G) Laughter }, fonttitle=\bfseries\small, breakable, sharp corners, boxrule=0.5pt, left=3pt, right=3pt, top=3pt, bottom=3pt]
{\scriptsize\ttfamily
CATEGORY: (G) Laughter .\par
Valid cue examples: smile / eye-crinkle / open-mouth amusement / chuckle / warm verbal ``haha'' or ``yeah'' (either channel suffices).\par
\texttt{cue\_timing} captures whether the laughter response starts during the user laugh window (or within 500ms of dynamic start) versus after window end.\par
\texttt{cue\_appropriateness} captures quality of produced laughter response without re-penalizing timing.\par
Wrong-cue cap: concerned/confused affect on a clearly funny moment.
}
\end{tcolorbox}

\begin{tcolorbox}[colback=magenta!3!white, colframe=magenta!50!black, title={(G) Nonverbal Backchanneling}, fonttitle=\bfseries\small, breakable, sharp corners, boxrule=0.5pt, left=3pt, right=3pt, top=3pt, bottom=3pt]
{\scriptsize\ttfamily
CATEGORY: (G) Nonverbal Backchanneling .\par
This category uses the unified backchannel rubric (shared with Verbal Backchanneling): visible cue OR short verbal acknowledgement counts as cue production.\par
Valid cue examples: nod, smile, brow-raise, posture shift, gaze settle, or brief verbal ``mhm'' / ``yeah'' ($\le$2 words).\par
\texttt{cue\_timing} distinguishes realtime listener feedback from turn-based late acknowledgment.\par
Wrong-cue cap: full turn-taking statement instead of brief acknowledgement.
}
\end{tcolorbox}

\begin{tcolorbox}[colback=magenta!3!white, colframe=magenta!50!black, title=(G) Verbal Backchanneling, fonttitle=\bfseries\small, breakable, sharp corners, boxrule=0.5pt, left=3pt, right=3pt, top=3pt, bottom=3pt]
{\scriptsize\ttfamily
CATEGORY: (G) Verbal Backchanneling .\par
Uses the same unified backchannel rubric as Nonverbal Backchanneling: either modality can satisfy \texttt{cue\_produced} when it functions as listener acknowledgement.\par
Audio-side transcript and segment timing are authoritative for short verbal acks.\par
\texttt{cue\_appropriateness} penalizes using the backchannel as turn-launch (long continuation after initial ack).
}
\end{tcolorbox}

\begin{tcolorbox}[colback=magenta!3!white, colframe=magenta!50!black, title=(G) Verbal Interruption, fonttitle=\bfseries\small, breakable, sharp corners, boxrule=0.5pt, left=3pt, right=3pt, top=3pt, bottom=3pt]
{\scriptsize\ttfamily
CATEGORY: (G) Verbal Interruption .\par
The target behavior is prompt yielding when the user interrupts: agent halts speech (or is already silent), shifts to listening, and stays yielded.\par
\texttt{cue\_timing} captures whether yielding starts in-window versus after prolonged talk-over.\par
\texttt{cue\_appropriateness} captures yield quality (listening shift, cooperative handoff, no immediate floor re-take).
}
\end{tcolorbox}

\begin{tcolorbox}[colback=magenta!3!white, colframe=magenta!50!black, title=(G) Emotion Matching; Face Emotion Display, fonttitle=\bfseries\small, breakable, sharp corners, boxrule=0.5pt, left=3pt, right=3pt, top=3pt, bottom=3pt]
{\scriptsize\ttfamily
CATEGORY: (G) Emotion Matching; Face Emotion Display.\par
Valid cue examples: affective signal in either channel aligned with user emotional tone (warmth, concern, excitement, etc.).\par
\texttt{cue\_timing} captures whether affective mirroring begins during the user-affect window versus late turn-based mirroring.\par
\texttt{cue\_appropriateness} captures semantic/emotional fit given a produced cue.\par
Wrong-cue cap: active contradiction (e.g., cheerful response to user distress).
}
\end{tcolorbox}

\begin{tcolorbox}[colback=magenta!3!white, colframe=magenta!50!black, title=(G) Turn-taking, fonttitle=\bfseries\small, breakable, sharp corners, boxrule=0.5pt, left=3pt, right=3pt, top=3pt, bottom=3pt]
{\scriptsize\ttfamily
CATEGORY: (G) Turn-taking .\par
This category has its own generation rubric in the current implementation (it is not folded into AV timing): the cue is overlap-tolerant boundary handling when the user re-enters.\par
Valid cue behavior: yield / halt / brief acknowledgment at boundary and no immediate floor re-take.\par
Wrong-cue cap: talking over user re-entry or restarting immediately after a short user pause.
}
\end{tcolorbox}

\newpage

\end{document}